\documentclass[graybox]{svmult}

\usepackage{type1cm}           
\usepackage{makeidx}           
\usepackage{graphicx}          
\usepackage{multicol}          
\usepackage[bottom]{footmisc}  

\usepackage{newtxtext}         %
\usepackage[varvw]{newtxmath}  
\usepackage{amsmath}

\usepackage{amssymb}

\usepackage{hyperref}
\usepackage{subcaption}
\usepackage{graphicx}
\usepackage{float}
\usepackage{algorithm}
\usepackage[noend]{algpseudocode}

\makeindex             

\newcommand{\bB}{\mathbf{B}}
\newcommand{\bD}{\mathbf{D}}
\newcommand{\br}{\mathbf{r}}
\newcommand{\bu}{\mathbf{u}}
\newcommand{\bU}{\mathbf{U}}
\newcommand{\bv}{\mathbf{v}}
\newcommand{\bV}{\mathbf{V}}
\newcommand{\bw}{\mathbf{w}}

\newcommand{\bx}{\mathbf{x}}
\newcommand{\bX}{\mathbf{X}}
\newcommand{\by}{\mathbf{y}}
\newcommand{\bY}{\mathbf{Y}}
\newcommand{\bz}{\mathbf{z}}
\newcommand{\bZ}{\mathbf{Z}}
\newcommand{\rhs}{\mathbf{f}}
\newcommand{\noise}{\mathbf{B}}
\newcommand{\R}{\mathbb{R}}
\newcommand{\N}{\mathbb{N}}
\newcommand{\abs}[1]{|#1|}
\newcommand{\norm}[1]{||#1||}
\newcommand{\inp}[1]{\langle #1 \rangle}
\newcommand{\zero}{\mathbf{0}}
\newcommand{\one}{\mathbf{1}}

\newcommand{\mbf}[1]{\boldsymbol{#1}}
\newcommand{\intkernel}{\phi}
\newcommand{\intkernele}{{\intkernel^{E}}}
\newcommand{\intkernela}{{\intkernel^{A}}}

\newcommand{\intkernelvare}{\varphi^{E}}
\newcommand{\intkernelvara}{\varphi^{A}}
\newcommand{\bintkernel}{{\mbf{\phi}}}

\newcommand{\bintkernelvar}{{\mbf{\varphi}}}

\newcommand{\rhsfo}{\mathbf{f}}
\newcommand{\force}{F}
\newcommand{\bbX}{\mathbb{X}}
\newcommand{\bbV}{\mathbb{V}}
\newcommand{\bbY}{\mathbb{Y}}
\newcommand{\bbZ}{\mathbb{Z}}
\newcommand{\bvar}{\mbf{\xi}}

\newcommand{\mE}{\mathcal{E}}
\newcommand{\hypspace}{\mathcal{H}}

\newcommand{\mHe}{\mathcal{H}_{{K}^E}}
\newcommand{\mHa}{\mathcal{H}_{{K}^A}}
\newcommand{\mK}{\mathcal{K}}

\newcommand{\mM}{\mathcal{M}}
\newcommand{\mS}{\mathcal{S}}
\newcommand{\mT}{\mathcal{T}}
\newcommand{\mY}{\mathcal{Y}}
\newcommand{\thetae}{{\theta^E}}
\newcommand{\thetaa}{{\theta^A}}

\newcommand{\bXi}{\boldsymbol\Xi}
\newcommand{\Id}{\mathbf{I}}
\newcommand{\bxi}{\boldsymbol\xi}
\newcommand{\bphi}{\boldsymbol\phi}
\newcommand{\bvarphi}{\boldsymbol\varphi}
\newcommand{\bhypspace}{\boldsymbol\hypspace}

\newcommand{\clof}[1]{\mathit{k}_{#1}}

\newcommand{\argmin}[1]{\underset{#1}{\operatorname{arg}\operatorname{min}}\;}

\begin{document}

\title*{Learning Collective Behaviors from Observation}
\author{Jinchao Feng and Ming Zhong}
\institute{Jinchao Feng \at Great Bay University, Dongguan, Guangdong, China, \email{jcfeng@gbu.edu.cn}
\and Ming Zhong \at Illinois Institute of Technology, Chicago, IL, USA \email{mzhong3@iit.edu}}
%
%
\maketitle
\abstract{We present a comprehensive examination of learning methodologies employed for the structural identification of dynamical systems. These techniques are designed to elucidate emergent phenomena within intricate systems of interacting agents. Our approach not only ensures theoretical convergence guarantees but also exhibits computational efficiency when handling high-dimensional observational data. The methods adeptly reconstruct both first- and second-order dynamical systems, accommodating observation and stochastic noise, intricate interaction rules, absent interaction features, and real-world observations in agent systems. The foundational aspect of our learning methodologies resides in the formulation of tailored loss functions using the variational inverse problem approach, inherently equipping our methods with dimension reduction capabilities.}

\section{Introduction}\label{sec:intro}
Data-driven modeling of governing differential equations for biological and physical systems from observations can be traced back to at least Gauss, Lagrange, and Laplace \cite{stigler1986}. Kepler's laws of basic planetary motion, Newton's universal law of gravitation, and Einstein's general relativity provide excellent examples for discovering physical laws to understand motion from data. The capture and subsequent description of such biological and physical laws have always been important research topics. We are particularly interested in data-driven modeling of collective behaviors, also known as self-organization. These behaviors can manifest as clustering~\cite{Krause2000SI}, flocking~\cite{CS02, VZ2012, Vicsek_model}, milling~\cite{Chuang2007}, swarming~\cite{TDOEKB2012}, and synchronization~\cite{Strogatz2000, OKeeffe2017}. These behaviors can be observed in spontaneous magnetization, Bose-Einstein condensation, molecular self-assembly, eusocial behaviors in insects, herd behavior, groupthink, and many others (see detailed references in \cite{BDT2017, tadmor2021, tadmor2023}). Proper mathematical modeling of these behaviors can provide predictions and control of large systems. Hence, we review a series of learning methods specifically designed to provide mathematical insight from data observed in systems demonstrating collective behaviors.

There are many different kinds of mathematical models for collective behaviors; we focus on the agent-based aspect of these models. Therefore, we assume that the observational data follows certain dynamical systems. To simplify our discussion, we use the following autonomous dynamical system:
\[
\dot\bY = \rhs(\bY), \quad \bY \in \R^D,
\]
where the time-dependent variable $\bY$ describes a certain state in the system (such as position, velocity, temperature, phase, opinion, or a combination of these), the function $\rhs: \R^D \rightarrow \R^D$ provides the change needed to update $\bY$. Given the observation of ${\bY(t)}_{t \in [0, T]}$, is it possible to identify $\rhs$ to explain the data within the given time interval $[0, T]$? Many system identification methods have been developed, such as learning parameterized systems via the maximum likelihood approach, which includes parameter estimation \cite{kasonga1990maximum, bishwal2011estimation, gomes2019parameter, chen2021maximum, sharrock2021parameter}, and nonparametric estimation of drift in the stochastic McKean-Vlasov equation \cite{genon2022inference, della2022lan, yao2022mean}, Sparse Identification of Nonlinear Dynamics (SINDy, \cite{sindy2016}), Weak SINDy that leverages the weak form of the differential equation and sparse parametric regression \cite{messenger2021learning, messenger2022cells}, Physics-Informed Neural Network (PINN, \cite{pinn2019}), Neural ODE \cite{neuralode2018}, and many more. However, in the case of high-dimensional data, i.e., $D \gg 1$, the identification task becomes computationally prohibitive and time-consuming.

To overcome the curse of dimensionality, we employ a special and effective dimension reduction technique by constructing our learning methods based on the unique structure of the right-hand side function, $\rhs$. This unique structure arises from the fact that these dynamical systems are used to model collective behaviors, from which global patterns can emerge via only local interaction between pairs of agents\footnote{Agents here can be referred to as particles, cells, bacteria, robots, UAV, etc.}. To simplify the discussion, we can focus on first-order systems given as follows:
\[
\dot\by_i = \frac{1}{N}\sum_{i' = 1, i' \neq i}^N\phi(\norm{\by_{i'} - \by_i})(\by_{i'} - \by_i), \quad i = 1, \cdots, N.
\]
Here, $\by_i \in \R^d$ describes the state of the $i^{th}$ agent, and $\phi:\R^+ \rightarrow \R$ gives the interaction law on how agent $i'$ influences the change of state for agent $i$. This system is a gradient flow of a certain system energy with rotation/permutation invariances and symmetry. Such a first-order system can be used to model opinion dynamics (formation of consensus), crystal structure, and other pattern formations in skin pigmentation \cite{SOinBio2003}. Using vector notation, i.e., letting
\[
\bY = \begin{bmatrix} \vdots \\ \by_i \\ \vdots \end{bmatrix} \in \R^{D = Nd} \quad \text{and} \quad \rhs_{\phi}(\bY) = \begin{bmatrix} \vdots \\ \frac{1}{N}\sum_{i' = 1, i' \neq i}^N\phi(\norm{\by_{i'} - \by_i})(\by_{i'} - \by_i) \\ \vdots \end{bmatrix} \in \R^D,
\]
we end up with the aforementioned autonomous dynamical system in the form of $\dot\bY = \rhs_{\phi}(\bY)$ with $\bY \in \R^D$ and $D = Nd \gg 1$. For example, even for a simple system with $10$ agents where $\by_i \in \R^2$, the system state variable lives in $D = 20$ dimensions. How to efficiently deal with such high dimensionality has become a crucial aspect of developing our learning algorithms \cite{lu2019nonparametric, ZHONG2020132542, pmlr-v139-maggioni21a, lu2021learning, FENG2022162, miller2023learning, zhong2021machine, feng2022learning, feng2023data}. Our methods are not only proven to have theoretical guarantees but are also verified through rigorous numerical testing. Even when learning from real observational data, our method can offer valuable insights into how the framework of collective behaviors can be used to explain complex systems.

The remaining sections of the paper are organized as follows: in Section \ref{sec:model}, we discuss the general model form used for our learning paradigm. In Section \ref{sec:learn}, we go through various learning scenarios, from first-order to second-order, from homogeneous agents to heterogeneous agents, and with Gaussian Process priors. We compare our methods to other well-established methods in Section \ref{sec:other_methods}. Finally, in Section \ref{sec:conclude}, we conclude our review with a highlight of several future directions.
\section{Model Equation}\label{sec:model}
In order to provide a more unified picture of modeling collective dynamics, we consider the following state variable $\by_i$ for the $i^{th}$ agent in a system of $N$ interacting agents. Moreover, $\by_i$ is expressed in the form of $\by_i = \begin{bmatrix} \bx_i \ \bv_i \ \zeta_i \end{bmatrix}$, where $\bx_i \in \R^d$ represents the position of agent $i$, $\bv_i \in \R^d$ is its corresponding velocity (hence $\bv_i = \dot\bx_i$), and $\zeta_i \in \R$ is an additional state that can be used to describe opinions, phases, excitation levels (e.g., towards light sources), emotions, etc. The coupled state variable $\by_i \in \R^{2d + 1}$ satisfies the following coupled dynamical system (for $i = 1, \cdots, N$):
\begin{equation}\label{eq:total_eqn}
\begin{cases}
    \dot\bx_i &= \bv_i \\
    m_i\dot\bv_i &= \force^{\bv}(\by_i) + \sum_{i' = 1, i' \neq i}^N\frac{1}{N_{\clof{i}}}\Big[\Phi^E_{\clof{i}, \clof{i'}}(\by_i, \by_{i'})\bw^E(\bx_i, \bx_{i'}) \\
    &\quad + \Phi^A_{\clof{i}, \clof{i'}}(\by_i, \by_{i'})\bw^A(\bv_i, \bv_{i'})\Big] \\
    \dot\zeta_i &= \force^{\zeta}(\by_i) + \sum_{i' = 1, i' \neq i}^N\frac{1}{N_{\clof{i}}}\Phi^{\zeta}_{\clof{i}, \clof{i'}}(\by_i, \by_{i'})(\zeta_{i'} - \zeta_i)
\end{cases}.
\end{equation}
Here, we consider the system to be partitioned into $K$ types, i.e., $C_{k_1} \cap C_{k_2} = \emptyset$ for $1 \le k_1 \neq k_2 = K$, $\cup_{k = 1}^K C_K = [N] = {1, \cdots, N}$, and $N_{k}$ is the set of agents in type $k$. Moreover, the types of the agents do not change over time\footnote{Such a restriction can be relaxed, as long as we have the type information for the agents at all times.}, and there is a type function $\clof{\cdot}:\N \rightarrow \N$, which returns the type index for an agent index. Furthermore, the external force $\force^{\bv}:\R^{2d + 1} \rightarrow \R^d$ gives the environmental effect on the velocity of the agent $i$, similarly $\force^{\zeta}:\R^{2d + 1} \rightarrow \R$ gives the environmental effect on $\xi_i$, $\Phi^E: \R^{4d + 2} \rightarrow \R$ gives the energy-based interaction, $\Phi^A: \R^{4d + 2} \rightarrow \R$ gives the alignment-based interaction, $\Phi^{\zeta}: \R^{4d + 2} \rightarrow \R$ gives the $\zeta$-based interaction, $\bw^E, \bw^A: \R^{2d} \rightarrow \R^d$ give the directions for $\Phi^E, \Phi^A$ respectively.
\begin{remark}
The inclusion of $(\bx_i, \bv_i)$ into the system also makes it a second-order system (second-order time derivative for $\bx_i$). We include the additional state $\zeta_i$ to provide a more realistic modeling of emergent behaviors. This formulation can also include first-order systems as special cases. When $\force^{\bv}(\by_i) = -\nu_i\bv_i$ (the usual friction) with $\nu_i \gg m_i$, $\Phi^A = 0$, and $\Phi^E$ depends on $(\bx_i, \zeta_i, \bx_{i'}, \zeta_{i'})$, the second-order systems become:
\[
\zero \approx -\bv_i + \frac{1}{\nu_i}\sum_{i' = 1, i' \neq i}^N\frac{1}{N_{\clof{i}}}\Phi^E_{\clof{i}, \clof{i'}}(\bx_i, \zeta_i, \bx_{i'}, \zeta_{i'})\bw^{E}(\bx_i, \bx_{i'}).
\]
This results in a first-order system of $\bx_i$ coupled with $\zeta_i$ (see the swarmalator model \cite{Strogatz2000, OKeeffe2017, hao2023}).

Our second-order coupled formulation \eqref{eq:total_eqn} can include a rather extensive list of behaviors, such as flocking with external potential \cite{shu2020}, anticipated flocking dynamics \cite{shu2021}, fish swarming dynamics \cite{FNSKE2018}, swarmalator dynamics (concurrent swarming and synchronization) \cite{Strogatz2000, OKeeffe2017, gerew2023concurrent, hao2023}, line alignment dynamics \cite{Greene_2023}, where the interaction kernel depends on pairwise state variables. It is possible to include more state variables; for example, we can also consider $(\bx_i, \bv_i, \zeta_i, \eta_i)$, where $\eta_i$ describes a different state than $\zeta_i$. However, learning such a system with four state variables adds mere technical complexity, and we will leave the details for future projects.
\end{remark}
Using the vector notation, we can obtain a simplified system denoted as $\dot\bY = \rhs_{\Phi^E, \Phi^A, \Phi^{\zeta}}(\bY)$, where $\bY = \begin{bmatrix} \cdots & \by_i^\top & \cdots \end{bmatrix}^\top \in \R^{D = N(2d + 1)}$ and
\[
\rhs(\bY) = \begin{bmatrix} \vdots \\ \bv_i \\\force^{\bv}(\by_i) + \sum_{i' = 1, i' \neq i}^N\frac{1}{N_{\clof{i}}}\Big[\Phi^E_{\clof{i}, \clof{i'}}(\by_i, \by_{i'})\bw^E(\bx_i, \bx_{i'}) \\
\quad + \Phi^A_{\clof{i}, \clof{i'}}(\by_i, \by_{i'})\bw^A(\bv_{i'} - \bv_i)\Big]\\  \force^{\zeta}(\by_i) + \sum_{i' = 1, i' \neq i}^N\frac{1}{N_{\clof{i}}}\Phi^{\zeta}_{\clof{i}, \clof{i'}}(\by_i, \by_{i'})(\zeta_{i'} - \zeta_i) \\ \vdots\end{bmatrix}  \in \R^{N(2d + 1)}.
\]

For such a coupled system, we seek to identify $\rhs = \rhs_{\{\Phi^E_{k_1, k_2}, \Phi^A_{k_1, k_2}, \Phi^{\zeta}_{k_1, k_2}\}_{k_1, k_2 = 1}^K}$ in terms of the interaction kernels $\{\Phi^E_{k_1, k_2}, \Phi^A_{k_1, k_2}, \Phi^{\zeta}_{k_1, k_2}\}_{k_1, k_2 = 1}^K$ from observations. With $\bY \in \R^{D = N(2d + 1)}$, the computational complexity of learning $\rhs$ increases exponentially. We will discuss the details of how to handle such high-dimensional learning in Section \ref{sec:learn}.

\section{Learning Framework}\label{sec:learn}
We are now prepared to discuss a unified learning framework developed for the efficient learning of such high-dimensional dynamical systems used to model collective behaviors. As mentioned in Section \ref{sec:intro}, we are interested in learning the dynamical system from observation in the form $\dot\by = \rhs(\by)$, where $\by \in \R^{D}$ with $D \gg 1$. However, due to the special structure in $\rhs$, i.e., $\rhs = \rhs_{\phi}$ where $\phi:\Omega \subset \R^{D'}$ with $D' \ll D$, we can exploit it and reduce the dimension for our learning framework. The framework focuses on constructing a suitable loss function designed for various $\varphi$'s from the observation data ${\bY(t), \dot\bY(t)}{t \in [0, T]}$ with $\bY(0) \sim \mu{\bY}$; i.e., the unified loss takes on the following form:
\[
\mE(\varphi) = \mathbb{E}_{\bY(0) \sim \mu_{\bY}}\Big[\int_{t = 0}^T\norm{\dot\bY(t) - \rhs_{\varphi}(\bY(t))}^2_{\mY}\, dt\Big],
\]
where the vector norm $\norm{\cdot}_{\mY}$ is designed for the special structure of $\bY$. We will begin our discussion from the simplest case: first-order systems, and then gradually extend the method to more complex scenarios, such as systems with multiple types of agents, stochastic noise, geometry-constrained dynamics, missing feature maps, coupled systems (for high-order systems), and learning with a Gaussian Process prior.

Before we dive into the details, we would like to provide a brief summary of the papers that contribute to various aspects of our learning framework. An initial framework on first- and second-order multi-species systems was developed and analyzed in \cite{lu2019nonparametric}, where the major focus is on cases with a fixed number of agents rather than the mean-field limit, i.e., $N \rightarrow \infty$, as discussed in \cite{BFHM17SI}.   A subsequent inquiry, directed towards the theoretical foundations of estimators applied to multi-species agents, was undertaken as detailed in \cite{lu2021learning}.  A numerical study of the steady-state behavior of our learned estimators was presented in \cite{ZHONG2020132542} with an extension to a two-dimensional interaction function $\phi$. A comprehensive analysis of second-order systems of heterogeneous agents with multi-dimensional $\phi$ was presented in \cite{miller2023learning}, compared to learning scenario examined in \cite{lu2021learning}, which involved first-order systems of heterogeneous agents characterised by one-dimensional interaction functions.  An extension to dynamics constrained on Riemannian manifolds was given in \cite{pmlr-v139-maggioni21a}, where the convergence rate similar to the one in Euclidean space was established by preserving the geometric structure in the learning. In \cite{FENG2022162}, a combination of feature map learning and dynamical system learning was presented to effectively reduce the high-dimensional interaction function $\Phi$ within the collective dynamics framework. An application to re-establish Newton's framework of the universal law of gravitation from NASA JPL's Horizon database (a highly accurate synthetic database of our solar system used in space exploration) was presented in \cite{zhong2021machine}, where our learning methods were able to provide relative errors at the scale of $10^{-8}$ to capture some of the general relativity effects within the limitation of the collective dynamics scheme. A learning framework with a Gaussian prior in order to reduce the number of learning samples needed and provide uncertainty quantification was presented in \cite{feng2022learning, feng2023data}.
\subsection{First Order}
We start with a simple first-order system to build our learning framework.  The first-order system which we consider is given as follows
\begin{equation}\label{eq:first_order}
\dot\bx_i = \frac{1}{N}\sum_{i' = 1, i' \neq i}^N\Phi(\bx_i, \bx_{i'})(\bx_{i'} - \bx_i), \quad i = 1, \cdots, N.
\end{equation}
Here, $\bx_i$ is the state variable of the $i^{th}$ agent, and $\Phi = \Phi^E: \R^{2d} \rightarrow \R$ gives the energy-based (short-range repulsion and long-range attraction; we omit the superscript $E$ here to streamline the notations) interaction. We only require the interaction kernel $\Phi$ to be symmetric, i.e., $\Phi(\bx_i, \bx_{i'}) = \Phi(\bx_{i'}, \bx_i)$. We shall further assume that $\Phi$ can be written in the following compositional form:
\begin{equation}\label{eq:feature_decomp}
\Phi(\bx_i, \bx_{i'}) = \phi(\bxi(\bx_i, \bx_{i})), \quad \phi: \R^{d_{\phi}} \rightarrow \R \quad \bxi: \R^{2d} \rightarrow \R^{d_{\phi}}.
\end{equation}
Here, $d_{\phi} \ll 2d$ is the number of feature variables for the interaction function, $\phi$ is the reduced interaction kernel, and $\bxi$ is the reduced interaction variable. In most of the collective behavior models, $\bxi$ is assumed to be the pairwise distance variable, i.e., 
\[
\bxi(\bx_i, \bx_{i'}) = \norm{\bx_{i'} - \bx_i}, \quad \norm{\bx} = \sqrt{\sum_{j = 1}^d \abs{(\bx)_j}^2}, \quad \forall \bx \in \R^d.
\]
\begin{figure}[H]
    \centering
    \begin{subfigure}[t]{0.48\textwidth}
        \centering
        \includegraphics[width = 0.9\textwidth]{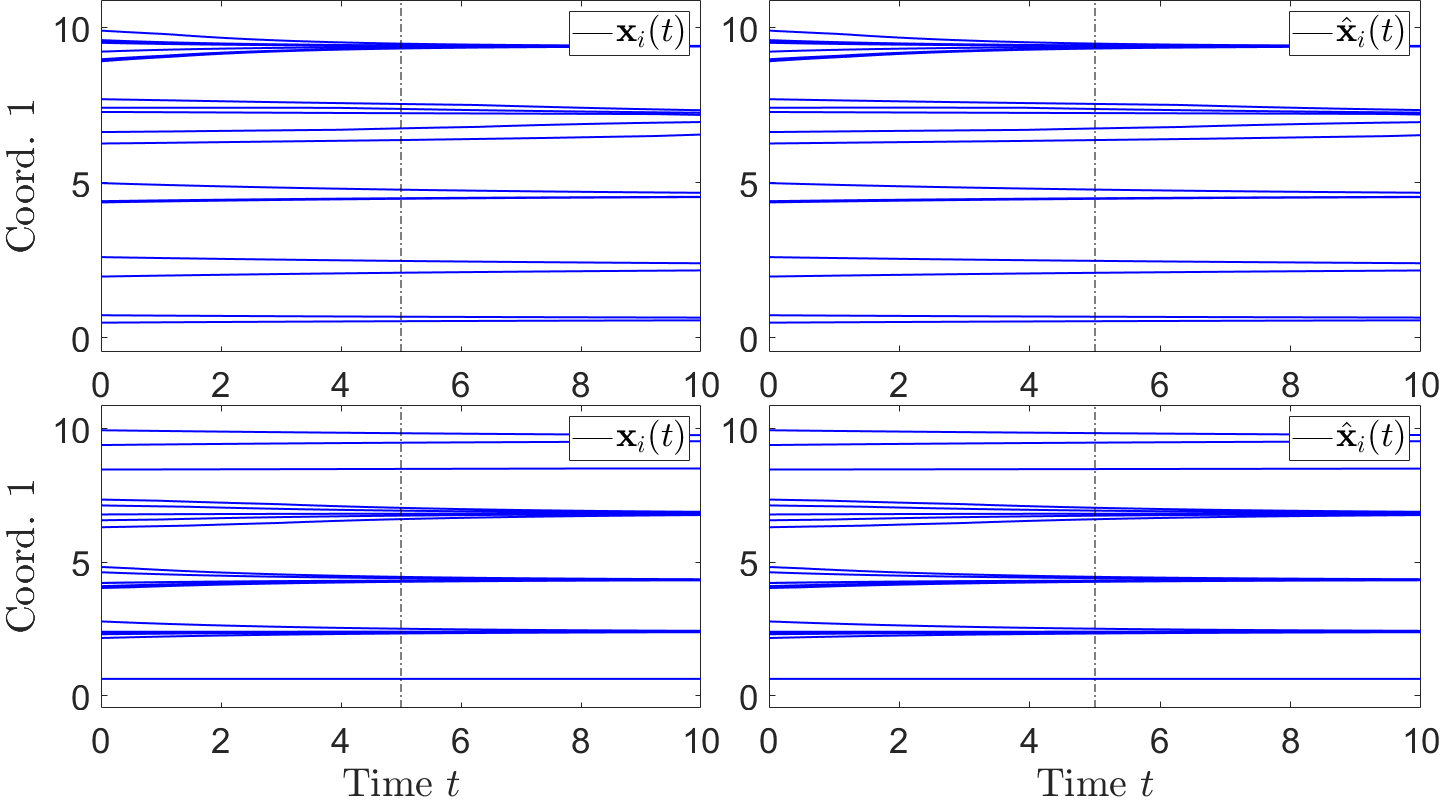}
        \caption{$\bX(t)$ vs $\hat\bX(t)$}
    \end{subfigure}%
    ~ 
    \begin{subfigure}[t]{0.48\textwidth}
        \centering
        \includegraphics[width = 0.9\textwidth]{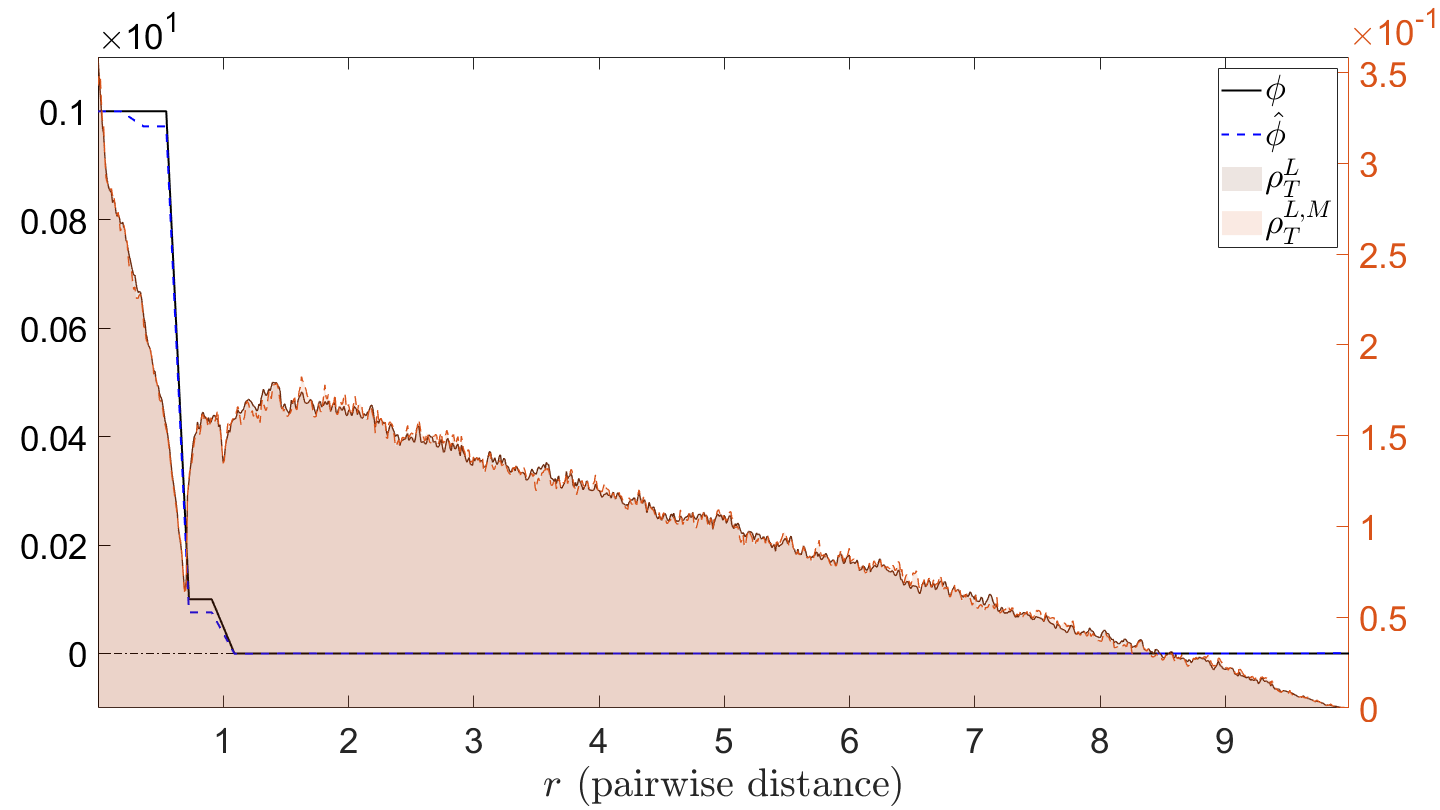}
        \caption{$\phi$ vs $\hat\phi$}
    \end{subfigure}
    \caption{Opinion Dynamics introduced in~\cite{Krause2000SI, MT2014}, with $N = 20$ agents and $\phi(r) = \chi_{[0, \frac{1}{\sqrt{2}})}(r) + 0.1*\chi_{[\frac{1}{\sqrt{2}}, 1]}(r)$, learned on $[0, 10]$, for other parameters see \cite{lu2019nonparametric}.}
\end{figure}
Hence we arrive at a simpler and more familiar form of \eqref{eq:first_order}
\begin{equation}\label{eq:first_order_simple}
\dot\bx_i = \frac{1}{N}\sum_{i' = 1, i' \neq i}^N\phi(\norm{\bx_{i'} - \bx_i})(\bx_{i'} - \bx_i), \quad i = 1, \cdots, N.
\end{equation}
\begin{remark}
\eqref{eq:first_order_simple} can be viewed as a gradient flow of the following form
\[
\dot\bx_i = \nabla_{\bx_i}\Big(\frac{1}{2N}\sum_{1 \le i \neq i \le N}U(\norm{\bx_{i'} - \bx_i})\Big), \quad i = 1, \cdots, N.
\]
Here $U: \R^+ \rightarrow \R$ is an energy potential that depends on pairwise distance.
\end{remark}
We are now ready to discuss the learning framework based on the $1$-dimensional $\phi$; such a learning method can be easily extended to $2$- or $3$-dimensional $\phi$ as presented in \cite{ZHONG2020132542, miller2023learning}. However, in order to efficiently handle the $2d$-dimensional $\Phi$, we leave the discussion in the feature map learning section. Furthermore, we consider observation data in this particular form, i.e., $\{\bx^m_i(t_l), \dot\bx^m_i(t_l)\}_{i, l, m = 1}^{N, L, M}$ for $0 = t_1 < t_2 < \cdots < t_L = T$, and $\bx_i^m(0)$s are i.i.d. samples of a certain probability distribution $\mu_0$; where $m$ indicates different initial conditions from the observation data. By using $m$, we bring back some level of data independence. We now can find the estimator for $\phi$ by minimizing the following loss functional over all test functions, $\varphi$, from a carefully designed convex and compact function/hypothesis space $\hypspace$, i.e.:
\begin{equation}\label{eq:lf_first_order_simple}
\mE_{L, M, \hypspace}(\varphi) = \frac{1}{LM}\sum_{l, m = 1}^{L, M}\norm{\dot\bX_{l}^m - \rhs_{\varphi}(\bX_l^m)}^2_{\mS}, \quad \varphi \in \hypspace.
\end{equation}
Here $\bX_l^m = \begin{bmatrix} \cdots & (\bx_i^m(t_l))^\top & \cdots \end{bmatrix}^\top \in \R^{D = Nd}$,
\[
\rhs_{\varphi}(\bX_l^m) = \begin{bmatrix}\vdots \\ \frac{1}{N}\sum_{i' = 1, i' \neq i}^N\varphi(\norm{\bx_{i'}^m(t_l) - \bx_{i}^m(t_l)})(\bx_{i'}^m(t_l) - \bx_i^m(t_l)) \\ \vdots \end{bmatrix} \in \R^{D},
\]
and the norm $\norm{\cdot}_{\mS}$ is defined as follows
\[
\norm{\bU - \bV}_{\mS}^2 = \frac{1}{N}\sum_{i = 1}^N \norm{\bu_i - \bv_i}^2_{\ell^2(\R^d)}, \quad \bU = \begin{bmatrix} \bu_1 \\ \vdots \\ \bu_i \\ \vdots \\ \bu_N\end{bmatrix}, \quad \bV = \begin{bmatrix} \bv_1 \\ \vdots \\ \bv_i \\ \vdots \\ \bv_N\end{bmatrix}, \quad \bu_i, \bv_i \in \R^d.
\]
Since $\hypspace$ is chosen to be convex and compact, the existence and uniqueness of the minimization are guaranteed.  We denote the the unique minimizer of $\mE_{L, M, \hypspace}$ as $\hat\phi_{L, M, \hypspace}$, i.e. $\hat\phi_{L, M, \hypspace} = \argmin{\varphi \in \hypspace}\mE_{L, M, \hypspace}(\varphi)$.  
\begin{remark}
By re-writing the dynamical system into $\dot\bX = \rhs_{\phi}(\bX)$, it seems we might be able to use the regression framework to learn $\rhs$, simply by letting the input be $\bX$ and the response be $\dot\bX$. However, due to the high dimensionality of $\bX$, this approach is computationally prohibitive, not to mention the lack of data independence from our observation data. But by exploiting the structure of $\rhs_{\phi}$, i.e., the dependence on $\phi$, the learning can be thought of as $1$-dimensional learning. However, it is not a regression learning since we do not have a direct observation of $\phi$ at various points; instead, we have a linear combination of $\phi$. Therefore, we need to cast it as a variational inverse problem to learn $\phi$.
\end{remark}
In order to demonstrate the convergence of $\hat\phi_{L, M, \hypspace}$ to $\phi$ as $M \rightarrow \infty$, we will need to consider the expected version of the loss functional, i.e.:
\[
\mE_{L, \hypspace}(\varphi) = \frac{1}{L}\mathbb{E}_{\bX_0 \sim \mu_0}\Big[\sum_{l}^{L}\norm{\dot\bX_l^m - \rhs_{\varphi}(\bX_l^m)}_{\mS}^2\Big].
\]
We denote the minimizer $\hat\phi_{L, \hypspace}$ from minimizing $\mE_{L, \hypspace}$ over the same hypothesis space $\hypspace$, i.e., $\hat\phi_{L, \hypspace} = \argmin{\varphi \in \hypspace}\mE_{L, \hypspace}(\varphi)$. Using the law of large numbers, $\hat\phi_{L, M, \hypspace} \rightarrow \hat\phi_{L, \hypspace}$ as $M \rightarrow \infty$. The only obstacle left is to bound the distance between $\hat\phi_{L, \hypspace}$ and $\phi$ due to the approximation power of $\hypspace$. Before we show the lemma needed to prove convergence, we need to define a weighted $L^2(\rho)$ norm where the weight function $\rho$ is given as follows:
\begin{equation}\label{eq:rho_def}
\rho(r) = \frac{1}{{N \choose 2}T}\int_{t = 0}^T\mathbb{E}_{\bX_0 \sim \mu_0}\Big[\sum_{i, i' = 1, i < i'}^N\delta_{r_{i, i'}(t)}(r) \, dt\Big].
\end{equation}
Here, $\br_{i, i'} = \bx_{i'} - \bx_i$ and $r_{i, i'} = \norm{\br_{i, i'}}$. $\rho$ is used to illustrate the distribution of pairwise distance data through the interaction of agents caused by the dynamics. Such data is also used to learn $\phi$; however, we do not have direct access to individual $\phi(r)$; rather, we have the indirect observation in terms of the linear combination of $\phi(r)$, i.e., $\sum_{j = 1, j \neq i}^N\phi(r_{i, j})\br_{i, j}$. With this definition of $\rho$, we can define our weighted $L^2(\rho)$ norm as follows:
\begin{equation}\label{eq:l2norm}
\norm{\phi_1(\cdot)\cdot - \phi_2(\cdot)\cdot}_{L^2(\rho)}^2 = \int_{r \in \text{supp}(\rho)} \abs{\phi_1(r) - \phi_2(r)}r\, d\rho(r),
\end{equation}
for any $\phi_1, \phi_2 \in L^2(\text{supp}(\rho), \rho)$.  Now we are ready to show the lemma needed to prove the convergence theorem.
\begin{definition}[Definition $3.1$ in \cite{lu2019nonparametric}]\label{def:cd}
The dynamical system given by \eqref{eq:first_order_simple} with IC sampled from $\mu_0$ on $\R^{D = Nd}$, satisfies the \textbf{coercivity condition} on a set $\hypspace$ if there exists a constant $C_{L, N, \hypspace}$ such that for all $\varphi \in \hypspace$ with $\varphi(\cdot)\cdot \in L^2(\rho)$, 
\[
C_{L, N, \hypspace}\norm{\varphi(\cdot)\cdot}^2_{L^2(\rho)} \le \frac{1}{LN}\sum_{i, l = 1}^{N, L}\mathbb{E}\norm{\frac{1}{N}\sum_{i' = 1, i' \neq i}^N \varphi(r_{i, i'}(t_l)\br_{i, i'}(t_l)}^2.
\]
\end{definition}
The coercivity condition, given by Definition \ref{def:cd}, basically states that minimizing $\mE_{L, \hypspace}$ would also minimize the distance between $\hat\phi_{L, \mathcal{H}}$ and $\phi$ under the weighted $L^2(\rho)$ norm, i.e.:
\[
C_{L, N, \hypspace}\norm{\phi(\cdot)\cdot - \varphi(\cdot)\cdot}^2_{L^2(\rho)} \le \mE_{L, \hypspace}(\varphi)
\]
To show this, one simply uses the fact that $\phi$ is the interaction kernel which gives the observation data, i.e. $\dot\bx_i = \frac{1}{N}\sum_{i' = 1, i' \neq i}^N\phi(r_{i, i'})\br_{i, i'}$.  With the coercivity condition established, we are ready to show the theorem \ref{thm:first_order_thm}.
\begin{theorem}[Theorem $3.1$ in \cite{lu2019nonparametric}]\label{thm:first_order_thm}
Assume that $\phi \in \mK_{R, S}$ (an admissible set $\mK_{R, S} = \{\phi \in C^1(\R_+): \text{supp}(\phi) \subset [0, R], \sup_{r \in [0, R]}\abs{\phi(r)} + \abs{\phi'(r)} \le S\}$ for some $R, S > 0$).  Let $\{\hypspace_n\}_n$ be a sequence of subspaces of $L^{\infty}([0, R])$, with $\text{dim}(\hypspace_n) \le c_0n$ and $\inf_{\varphi \in \hypspace_n}\norm{\varphi - \phi}_{L^{\infty}([0, R])} \le c_1n^{-s}$, for some constants $c_0, c_1, s > 0$.  Assume that the coercivity condition holds on $\hypspace \coloneqq \overline{\cup_{n = 1}^{\infty}\hypspace_n}$.  Such a sequence exists, for example, if $\phi$ is $s$-H\"{o}lder regular, and can be chosen so that $\hypspace$ is compact in $L^2(\rho)$.  Choose $n_* = (\frac{M}{\log(M)})^{\frac{1}{2s + 1}}$.  Then, there exists a constant $C = C(c_0, c_1, R, S)$ such that
\[
\mathbb{E}\Big[\norm{\hat\phi_{L, M, \hypspace_{n_*}}(\cdot)\cdot - \phi(\cdot)\cdot}_{L^2(\rho)}\Big] \le \frac{C}{C_{L, N, \hypspace}}(\frac{M}{\log(M)})^{\frac{s}{2s + 1}}.
\]
\end{theorem}
Our learning rate is as optimal as if it were learned from the regression setting, and the proof is an elegant combination of the proof presented in \cite{BFHM17SI} and \cite{cucker2002mathematical}. For a more detailed discussion and the actual proof, see \cite{lu2019nonparametric} and its supplementary information. Furthermore, our method is also robust against observation noise, as shown in Figure $8$ in \cite{lu2019nonparametric}.

When we choose a basis for $\hypspace_n$, i.e., $\hypspace_n = \text{span}\{\psi_1, \cdots, \psi_n\}$ and let $\varphi = \sum_{\eta = 1}^n \alpha_{\eta}\psi_{\eta}$, then the aforementioned minimization problem can be re-written as a linear system $A\vec{\alpha} = \vec{b}$, where $\vec{\alpha} = \begin{bmatrix} \alpha_1 & \cdots & \alpha_n \end{bmatrix}^\top$, $A \in \R^{n \times n}$ with
\[
A_{i, j} = \frac{1}{LM}\sum_{l, m = 1}^{L, M}<\rhs_{\psi_i}(\bX^m_l), \rhs_{\psi_j}(\bX^m_l)>_{\mS}, \quad <\bU, \bU>_{\mS} = \norm{\bU}_{\mS}^2,
\]
and $\vec{b} \in \R^{n}$ with
\[
\vec{b}_i = \frac{1}{LM}\sum_{l, m = 1}^{L, M}<\rhs_{\psi_i}(\bX^m_l), \dot\bX^m_l>_{\mS}.
\]
Moreover, we have a theoretical guarantee for the well-conditioning of the system; hence, it is computationally tractable to solve $A\vec{\alpha} = \vec{b}$. Algorithm \ref{alg:main} shows the pseudo-code for implementing the learning framework.
\begin{algorithm}[H]
\caption{Learning $\phi$ from Observations of First Order System}\label{alg:main}
\begin{algorithmic}[1]
\State Input: $\{\bx_i^{m}(t_l)$ and/or $\dot\bx_i^{m}(t_l)$.
\State Output: estimators for the interaction kernels.
\State Find out the maximum/minimum interaction radii $R_{\min}$, $\R_{\max}$.
\State Construct the basis, $\{\psi_{\eta}\}_{\eta = 1}^n$ on $[R_{\min}, R_{\max}]$.
\State Assemble $A\vec{\alpha} = \vec{b}$ (in parallel). 
\State Solve for $\vec{\alpha}$.
\State Assemble $\hat\phi(r) = \sum_{\eta = 1}^{n}\alpha_{\eta}\psi_{\eta}(r)$.
\end{algorithmic}
\end{algorithm}
Furthermore, for massive data sets, one can use parallelization to assemble $A$ and $\vec{b}$, which makes the overall process (near-)linear.  See \cite{lu2019nonparametric} for the detailed description of the algorithm, \url{https://github.com/mingjzhong/LearningDynamics} for the software package, and \url{https://youtu.be/yc-AIAEtGDc?si=cv9TckRX_grMCOt8} for on how to use the software package.
\subsubsection{Other System Identification Methods}\label{sec:other_methods} %
There are many different methods for learning dynamical systems, such as using the parametric structure of the right-hand side with Bayesian inference, leveraging the sparse structure of the right-hand side (SINDy \cite{sindy2016}), employing random features to approximate the interaction kernel \cite{liu2023random}, and using neural networks to estimate the right-hand side (PINN \cite{pinn2019} and NeuralODE \cite{neuralode2018}). We will discuss SINDy and the Neural Network approach and make a direct comparison of these two methods with ours.

SINDy (Sparse Identification of Nonlinear Dynamics) is a data-driven identification method for dynamical systems. It assumes that the dynamical system $\dot\bY = \rhs(\bY)$ can be approximated using the following form:
\[
\rhs(\bY) \approx a_1\psi_1(\bY) + \cdots + a_n\psi_n(\bY), \quad \bY = \begin{bmatrix} (\bY)_1 \\ \vdots \\ (\bY)_D \end{bmatrix} \in \R^D.
\]
$\psi_i$ is applied component wise to $\bY$, i.e.
\[
\psi_i(\bY) = \begin{bmatrix} \psi_i(y_1) \\ \vdots \\ \psi_i(y_D) \end{bmatrix}, \quad \bY = \begin{bmatrix} y_1 \\ \vdots \\ y_D \end{bmatrix}.
\]
Furthermore, $\{\psi_1, \cdots, \psi_n\}$ is a set of basis functions from a predetermined dictionary; for example, they can be a set of polynomials, sine/cosine, and/or negative power polynomials.  Given the observation data $\{\bY(t_1), \cdots, \bY(t_L)\}$, we assemble the matrix:
\[
\mathbf{Y} = \begin{bmatrix} \bY^\top(t_1) \\ \vdots \\ \bY^\top(t_L) \end{bmatrix} = \begin{bmatrix} y_1(t_1) & y_2(t_1) & \cdots & y_D(t_1) \\ y_1(t_2) & y_2(t_2) & \cdots & y_D(t_2) \\ \vdots & \vdots & \ddots & \vdots \\ y_1(t_L) & y_2(t_L) & \cdots & y_D(t_L)\end{bmatrix} \in \R^{L\times D}.
\]
We construct a dictionary (or library) $\Theta(\mathbf{Y})$ of nonlinear candidate functions of $\mathbf{Y}$, i.e.:
\[
\Theta(\mathbf{Y}) = \begin{bmatrix} \vert & \vert & & \vert  \\ \psi_1(\mathbf{Y}) & \psi_2(\mathbf{Y}) & \cdots & \psi_n(\mathbf{Y}) \\ \vert & \vert & & \vert\end{bmatrix} \in \R^{L\times nD}.
\]
Again $\psi_i(\mathbf{Y})$ is applied component wise, i.e.:
\[
\psi_i(\mathbf{Y}) = \begin{bmatrix}  \psi_i(y_1(t_1)) & \psi_i(y_2(t_1)) & \cdots & \psi_i(y_D(t_1)) \\ \psi_i(y_1(t_2)) & \psi_i(y_2(t_2)) & \cdots & \psi_i(y_D(t_2)) \\ \vdots & \vdots & \ddots & \vdots \\ \psi_i(y_1(t_L)) & \psi_i(y_2(t_L)) & \cdots & \psi_i(y_D(t_L))\end{bmatrix}.
\]
The set of basis functions can be $\{1, y, y^2, y^3, \cdots, \sin(y), \cos(y), \cdots\}$, depending on prior knowledge or computational capacity.  We write the dynamical system in the new form:
\[
\dot{\mathbf{Y}} = \Theta(\mathbf{Y})\Xi, \quad \Xi = \begin{bmatrix} \bxi_1 & \cdots & \bxi_D \end{bmatrix}, \quad \bxi_i \in \R^{nD \times D}.
\]
SINDy assumes that the expansion of $\Theta(\mathbf{Y})$ is the same for all time, it simplifies the structure of $\Xi$ down to
\[
\Xi = \begin{bmatrix} \bxi & \bxi & \cdots & \bxi \end{bmatrix} = \bxi \cdot \one, \quad \bxi \in \R^{nD \times 1} \quad \text{and} \quad \one = \begin{bmatrix} 1 & 1 & \cdots & 1 \end{bmatrix} \in \R^{1 \times D}.
\]
Furthermore, SINDy assumes that $\rhs(\bY(t))$ admits a spare representation in $\Theta(\mathbf{Y})$.  Moreover, it finds a parsimonious model by performing least squares regression with sparsity-promoting regularization, i.e.:
\[
\hat\bxi = \argmin{\bxi \in \R^{nD\times 1}}\{\norm{\dot{\mathbf{Y}} - \Theta(\mathbf{Y})(\bxi\cdot\one)}_{\text{Frob}}^2 + \lambda\norm{\bxi}_1\}.
\]
The SINDy approach has been shown to be effective for observational data from various kinds of dynamics, especially for the Lorenz system \cite{sindy2016, Shea2020SINDyBVPSI, Kaheman2020SINDyPIAR}. The assumption that the right-hand side function can be approximated well by a set of predetermined dictionaries opens up a question on how to optimally choose such a dictionary. In the case of collective dynamics, a direct application is ineffective. Although the form of collective dynamics can be written as polynomials in components of $\by_i$, the coefficients change all the time, and they rarely assume a sparse representation in terms of the coordinate systems (or any coordinate system). However, they might have a sparse representation in terms of the pairwise variables. In order to use it for collective dynamics, one has to re-work the approximation scheme; see \cite{weaksindy} for an example.

We can also use neural network structures to solve and infer information about dynamical systems. Given the observation data $\{\bY(t_l), \dot\bY(t_l)\}_{l = 1}^L$, we find the estimator to $\rhs$ as in $\dot\bY = \rhs(\bY)$ from minimizing the following loss functional:
\[
\text{Loss}(\tilde\rhs) = \frac{1}{L}\sum_{l = 1}^L\norm{\dot\bY(t_l) - \tilde\rhs(\bY(t_l))}_{\ell^2(\R^{D})}^2, \quad \tilde\rhs \in \mathcal{H}.
\]
Here $\tilde\rhs: \R^D \rightarrow \R^D$ and $\mathcal{H}$ is a set of neural networks of the same depth, the same neurons on each hidden layer, and the same activation function on each hidden layer.  The neural network solution is
\[
\hat\rhs = \argmin{\tilde\rhs \in \mathcal{H}}\text{Loss}(\tilde\rhs).
\]
Neural network approximation has been shown to be effective for high-dimensional function estimation. However, the right-hand side $\rhs(\bY)$ might have a special structure that a usual neural network might fail to capture. In this case, NeuralODE \cite{neuralode2018}, which uses the ODE solver and Recurrent Neural Network structure for training, can be employed to capture special properties of the ODE system. However, in the case of collective dynamics, when $\rhs$ on $\bY$ is learned directly, we limit ourselves to a fixed system (fixed number of agents). Recall that $\bY$ is a concatenation of all system state variables. In other words, if we switch to a system where the number of agents is different from the training data, we have to re-train. PINN (Physics Informed Neural Networks \cite{pinn2019}), on the other hand, may be adopted for such special needs, as it can learn both the right-hand side and the ODE solution together. A comprehensive study on the comparison between vanilla SINDy and NeuralODE was conducted in \cite{lu2021learning}; we are planning a more comprehensive comparison study where both the SINDy and NeuralODE are reformulated to conform to the collective dynamics models.

Another similar method is introduced as a random feature learning in \cite{liu2023random}, where the loss function combines both the observation loss and regularization on $\varphi \in \hypspace$, i.e.
\[
\mE_{\hypspace}(\varphi) = \mathbb{E}_{\bY_0 \sim \mu_{\bY}}\Big[\frac{1}{T}\int_{t = 0}^T\norm{\dot\bY_t - \rhs_{\varphi}(\bY_t)}_{\mS}^2 + \lambda J(\varphi)\Big].
\]
Here $\lambda$ is a regularization parameter, $J:\hypspace \rightarrow \R^+$ is a regularizing function on $\varphi$, and the function/hypothesis space $\hypspace$ uses random features as the basis.  The addition of regularization is also presented in \cite{feng2023data}.
\subsection{Heterogeneous Agents}
It is natural to consider heterogeneous agents, i.e. agents of different types, for there are many dynamics that have multiple types of agents involved, e.g. leader-follower, predator-prey, pedestrian-vehicle, etc.  We consider the scenario that the system of $N$ agents is partitioned into $K$ types, i.e. $C_{k_1} \cap C_{k_2} = \emptyset$ for $1 \le k_1 \neq k_2 = K$ and $\cup_{k = 1}^K C_K = [N] = \{1, \cdots, N\}$.  Moreover, the type of the agents does not change in time\footnote{Such restriction can be relaxed, as long as we know the type information for the agents at all time.}, and there is a type function $\clof{\cdot}:\N \rightarrow \N$, which returns the type index for an agent index.  
\begin{figure}[H]
    \centering
    \begin{subfigure}[t]{0.48\textwidth}
        \centering
        \includegraphics[width = 0.9\textwidth]{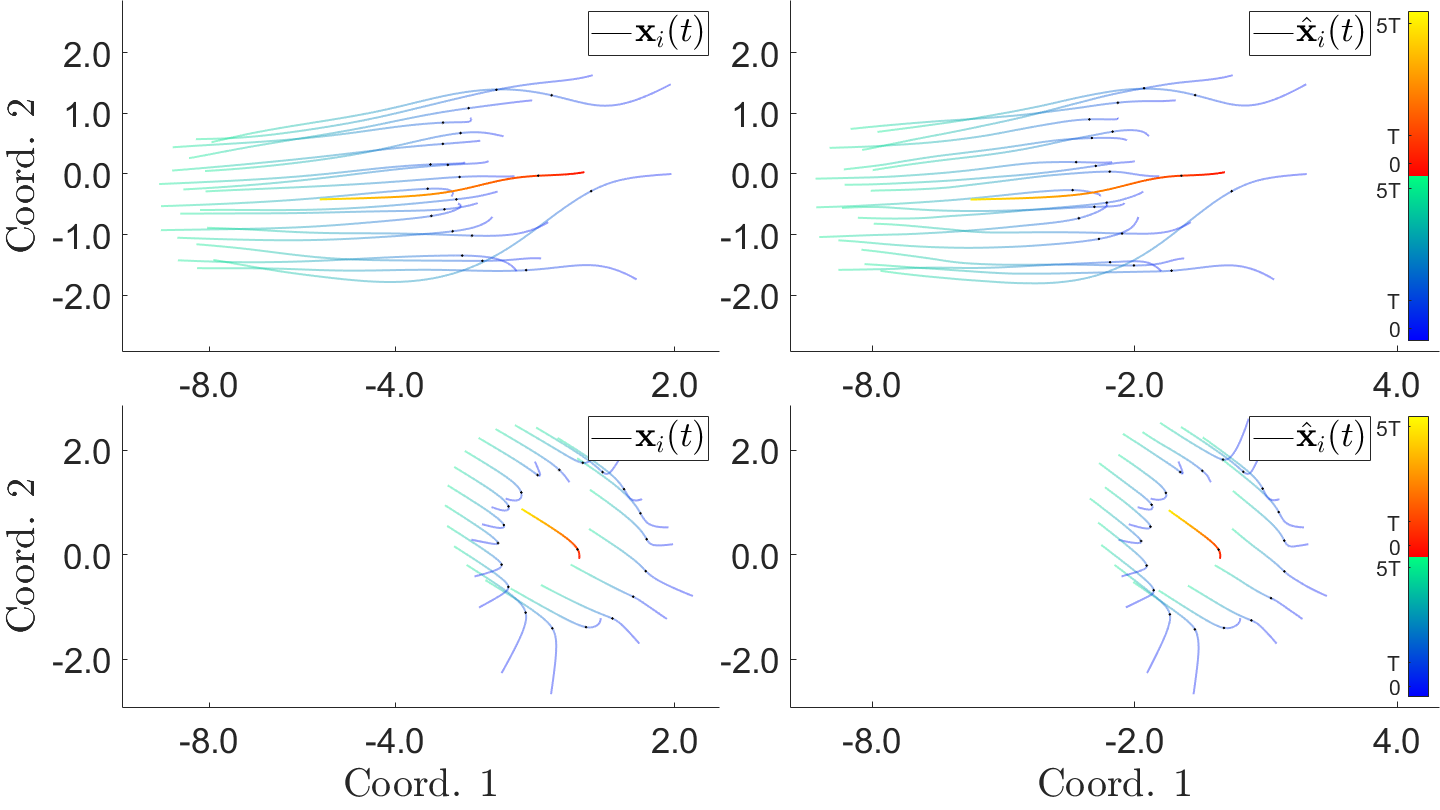}
        \caption{$\bX(t)$ vs $\hat\bX(t)$}
    \end{subfigure}%
    ~ 
    \begin{subfigure}[t]{0.48\textwidth}
        \centering
        \includegraphics[width = 0.9\textwidth]{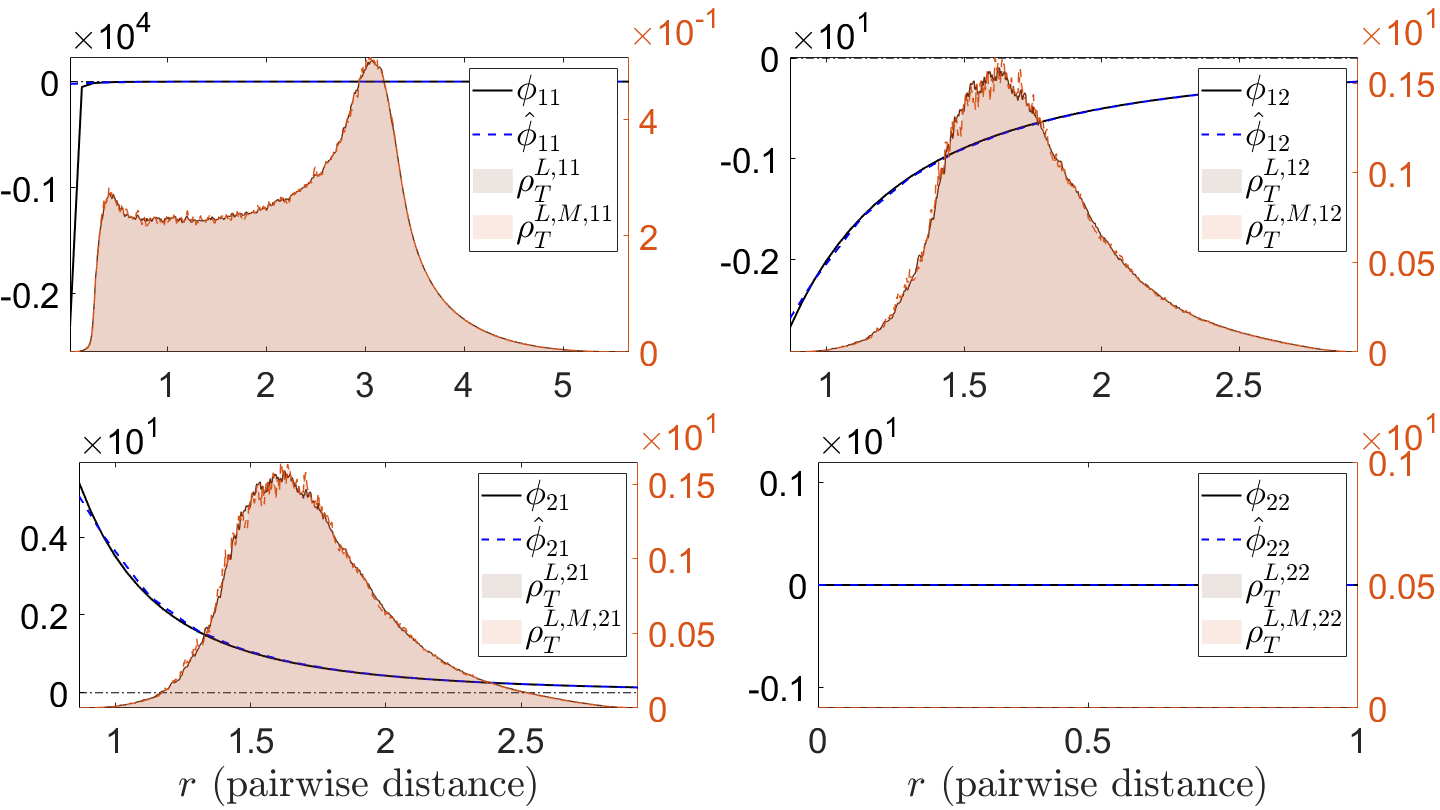}
        \caption{$\phi_{k_1, k_2}$ vs $\hat\phi_{k_1, k_2}$}
    \end{subfigure}
    \caption{Predator-Preys Dynamics introduced in~\cite{CK2013}, with $1: \text{prey}$, $2: \text{predator}$, $N_{1} = 19$ preys, $N_2 = 1$ predator, learned on $[0, 5]$, for other parameters see \cite{lu2019nonparametric}.}
\end{figure}
Then we can consider a simple extension of \eqref{eq:first_order_simple} as follows
\begin{equation}\label{eq:first_order_multi}
\dot\bx_i = \sum_{i' = 1, i' \neq i}^N\frac{1}{N_{\clof{i}}}\phi_{\clof{i}, \clof{i'}}(\norm{\bx_{i'} - \bx_i})(\bx_{i'} - \bx_i), \quad i = 1, \cdots, N.
\end{equation}
Now, the interaction kernel $\phi_{k_1, k_2}$ not only depends on the pairwise distance variable $r$ but is also different for agents of different types.  Given the same observation data, i.e. $\{\bx^m_i(t_l), \dot\bx^m_i(t_l)\}_{i, l, m = 1}^{N, L, M}$, we will learn a total of $K^2$ interaction kernels, i.e. $\bphi = \{\phi_{k_1, k_2}\}_{k_1, k_2 = 1}^{K}$, all together by minimizing a slightly updated loss functional, 
\begin{equation}\label{eq:lf_first_order_multi}
\mE^{\text{Mul}}_{L, M, \bhypspace}(\bvarphi) = \frac{1}{LM}\sum_{l, m = 1}^{L, M}\norm{\dot\bX_l^m - \rhs_{\bvarphi}(\bX_l^m)}_{\mS'}^2,
\end{equation}
Here $\bvarphi = \{\varphi_{k_1, k_2} \in \hypspace_{k_1, k_2}\}_{k_1, k_2 = 1}^K$, the direct sum space $\bhypspace = \oplus_{k_1, k_2 = 1}^K\mathcal{H}_{k_1, k_2}$, $\bX_l^m$ is the usual vector notation and 
\[
\rhs_{\bvarphi}(\bX_l^m) = \begin{bmatrix} \vdots \\ \sum_{i' = 1, i' \neq i}^N\frac{1}{N_{\clof{i}}}\varphi_{\clof{i}, \clof{i'}}(\norm{\bx_{i'}^m(t_l) - \bx_i^m(t_l)})(\bx_{i'}^m(t_l) - \bx_i^m(t_l)) \\ \vdots \end{bmatrix}.
\]
The norm, $\norm{\cdot}_{\mS'}$, is changed slightly as
\[
\norm{\bU - \bV}_{\mS'}^2 = \sum_{i = 1}^N\frac{1}{N_{\clof{i}}}\norm{\bu_i - \bv_i}^2_{\ell^2(\R^d)}, \quad \bU, \bV \in \R^{D = Nd}.
\]
The main difficulty in learning multiple $\phi_{k_1, k_2}$ at the same time is that there is no way to separate the trajectory data into $K^2$ groups. However, we are able to capture each individual $\phi_{k_1, k_2}$ differently from the same set of trajectory data. For proof of convergence, it follows similarly from the proof presented in \cite{lu2019nonparametric, lu2021learning}.
\subsection{Stochastic Noise}
Noise comes in various forms; it might appear in the observation data due to imperfect measurements, or it might manifest in the model as a stochastic noise to reflect the randomness in decision-making of these interacting agents. We consider the second scenario where a stochastic noise term is added to the first-order system \eqref{eq:first_order_multi}. It is given as follows (recall that $\bx_i \in \R^d$ represents the state of the $i^{th}$ agent):
\begin{equation}\label{eq:first_order_stoch}
d\bx_i = \Big(\sum_{i' = 1, i' \neq i}^N\frac{1}{N_{\clof{i}}}\phi_{\clof{i}, \clof{i'}}(\norm{\bx_{i'} - \bx_i})(\bx_{i'} - \bx_i)\Big)dt + d\noise_i,
\end{equation}
for $i = 1, \cdots, N$.  Here $d\noise_i$ is a Brownian motion with a state-dependent symmetric and positive-definite covariance matrix $\bXi_i(\bx)$ ($\bXi: \R^d \rightarrow \R^{d \times d}$).  Using vector-notation, i.e. let
\[
\rhs_{\bvarphi}(\bX_l^m) = \begin{bmatrix} \vdots \\ \sum_{i' = 1, i' \neq i}^N\frac{1}{N_{\clof{i}}}\varphi_{\clof{i}, \clof{i'}}(\norm{\bx_{i'}^m(t_l) - \bx_i^m(t_l)})(\bx_{i'}^m(t_l) - \bx_i^m(t_l)) \\ \vdots \end{bmatrix},
\]
where $\bvarphi = \{\varphi_{k_1, k_2}\}_{k_1, k_2 = 1}^K$, and
\[
\bD(\bX) = \begin{bmatrix}\bXi_1(\bx_1) & 0 & \cdots & 0 \\ 0 & \bXi_2(\bx_2) & \cdots & 0 \\ \vdots & \vdots & \ddots & \vdots \\ 0 & 0 &  \cdots & \bXi_N(\bx_N) \end{bmatrix} \quad \text{and} \quad \bB_t = \begin{bmatrix} \noise_1(t) \\ \vdots \\ \noise_N(t) \end{bmatrix} \in \R^{D \times D},
\]
The observation data is given in a slightly different form, i.e. $\{\bX_t, d\bX_t\}_{t \in [0, T]}$ with $\bX_0 \sim \mu$ is given, we find the minimizer from the following loss:
\[
\begin{aligned}
\mE_{\hypspace}^{\text{Sto}}(\varphi) &= \mathbb{E}_{\bX_0 \sim \mu_0}\Big[\frac{1}{2T}\int_{t = 0}^T<\rhs_{\bvarphi}, \bD^{-2}(\bX_t)\rhs_{\bvarphi}(\bX_t)> \, dt \\
&\qquad - \frac{1}{T}\int_{t = 0}^T<\rhs_{\bvarphi}(\bX_t), \bD^{-2}(\bX_t)d\bX_t>\Big].
\end{aligned}
\]
In actual applications, we will only be given snapshots of the states, i.e., ${\bX_l^m}{l, m = 1}^{L, M}$. We must approximate both $d\bX_l^m$ and the time integral in the loss. An initial study of the algorithm and its convergence was investigated in \cite{guo2024SDELearn}, where the automatic learning of the noise $\sigma$ and the drift term $\rhs_{\bphi}$ is combined.
\begin{remark}
Upon contemplation of a system comprising homogeneous agents, where the covariance matrix for each $\noise_i$ adopts the structure $\bXi_i(\bx) = \sigma\Id_{d \times d}$ with $\sigma > 0$ as a constant shared across all $\noise_i$, the loss function $\mE_{\hypspace}^{\text{Sto}}$ can be rendered in a simplified form,
\[
\mE_{\hypspace}^{\text{Sto}}(\varphi) = \frac{1}{2\sigma^2T}\mathbb{E}_{\bX_0 \sim \mu_0}\Big[\int_{t = 0}^T(\norm{\rhs_{\varphi}}^2dt - 2<\rhs_{\varphi}(\bX_t), d\bX_t>)\Big].
\]
The learning scenario characterized by a constant noise level devoid of correlation with other components or agents, particularly in the context of homogeneous agents, has been thoroughly investigated and demonstrated to exhibit superior convergence properties as documented in \cite{lu2022}. Notably, the recently introduced loss function in \cite{guo2024SDELearn} has expanded the scope of learning capabilities, enabling the accommodation of state-dependent and correlated noise across heterogeneous types of agents.  Both of the aforementioned loss functions draw inspiration from the Girsanov theorem. Nevertheless, the novel loss function additionally accounts for the correlated noise scenario, as expounded in Theorem $7.4$ within the framework presented in \cite{Sarkka_Solin_2019}.
\end{remark}
\subsection{Riemannian Geometry Constraints}
The state variable, i.e. $\bx_i \in \R^d$, might be living on a low-dimensional manifold as $d$ gets large. We consider the case when $\bx_i \in \mM \subset \R^d$, where $\mM$ is a Riemannian manifold naturally embedded in $\R^d$. We further assume that we know enough information about the manifold, i.e., the pair $(\mM, g)$, with $g$ being the Riemannian metric, is given to us. We also consider that the updated first-order model is as follows:
\begin{equation}
\bv_i = \sum_{i' = 1, i' \neq i}^N\frac{1}{N_{\clof{i}}}\phi_{\clof{i}, \clof{i'}}(\norm{\bx_{i'} - \bx_i})\bw^{\bx}(\bx_i, \bx_{i'}), \quad i = 1, \cdots, N.
\end{equation}
Here $\bv_i = \dot\bx_i \in \mT_{\bx_i}\mM$ (where $\mT_{\bx}\mM$ is the tangent space of $\mM$ at $\bx$), and $\bw^{\bx}(\bx_i, \bx_{i'})$ gives the unit tangent direction on the geodesic from $\bx_i$ to $\bx_{i'}$ if $\bx_{i'}$ is not in the cut locus set of $\bx_i$, otherwise $\bw^{\bx}(\bx_i, \bx_{i'}) = \mathbf{0}$. We also assume that each $\phi_{k_1, k_2}$ is defined on $[0, R]$, where $R$ is sufficiently small so that length-minimizing geodesics exist uniquely. Then such a first-order model is a well-defined gradient flow model of a sum of pairwise potential energy. To respect the geometry, we update the loss function so that it has the information about the Riemannian metric, 
\[
\mE^{\text{Rie}}_{L, M, \bhypspace}(\bvarphi) = \frac{1}{LM}\sum_{l, m = 1}^{L, M}\norm{\dot\bX_l^m - \rhs_{\bvarphi}(\bX_l^m)}_{g}^2,
\]
where the new norm, $\norm{\cdot}_{g}$, is given as
\[
\norm{\dot\bX_l^m - \rhs_{\bvarphi}(\bX_l^m)}_g^2 = \sum_{i = 1}^N\frac{1}{N_{\clof{i}}}\norm{\dot\bx_{i, l}^m - \sum_{i' = 1, i' \neq i}^N\frac{1}{N_{\clof{i'}}}\varphi_{\clof{i}, \clof{i'}}(r_{i, i', l}^m)\bw^m_{i, i', l}}_{\mT_{\bx_{i, l}^m}\mM}^2.
\]
Here $\bx_{i, l}^m = \bx_i^m(t_l)$, $r_{i, i', l}^m = |\bx_{i'}^m(t_l) - \bx_{i}^m(t_l)|$, and $\bw^m_{i, i', l} = \bw^{\bx}(\bx_{i, l}^m, \bx_{i', l}^m)$. With this new norm together with an updated coercivity condition respecting the geometric structure of the data, we are able to preserve the convergence rate as if the data is in Euclidean space, see \cite{pmlr-v139-maggioni21a} for details.
\subsection{Feature Map Learning}
Although there are many interacting-agent systems that demonstrate various complex emergent behaviors that are only involved with the interaction kernel functions depending on the pairwise distance as we described above, in real-life applications, the interactions may depend on some other unknown but a small number of variables (e.g. pairwise distances/angles of velocities, pairwise differences of phases).  To consider this needed complexity, we consider the interacting-agent systems of $N$ agents governed by the equations in the form \eqref{eq:first_order}, i.e.,
\begin{equation}
\dot\bx_i(t) = \frac{1}{N}\sum_{i' = 1}^N\Phi(\bx_i(t), \bx_{i'}(t))(\bx_{i'}(t) - \bx_i(t))\,,
\end{equation}
for $i = 1, \ldots, N$, $t \in [0, T]$, $\bx_i \in \R^d$ is a state vector, 
and $\Phi:\R^d\times\R^d \rightarrow \R$ is the \textbf{interaction kernel}, governing how the state of agent $i'$ influences the state of agent $i$. Note that while the state space of the system is $dN$-dimensional, the interaction kernel is a function of $2d$ dimensions. Moreover, we consider the case that the interaction kernel $\Phi$ is a function that depends on a smaller number of natural variables $\bvar\in\mathbb{R}^{d'}$, with $d'\ll 2d$, which are functions of pairs $(\bx_i,\bx_{i'})\in\R^d\times\R^d$. 
\begin{figure}[H]
    \centering
    \begin{subfigure}[t]{0.48\textwidth}
        \centering
        \includegraphics[width = 0.9\textwidth]{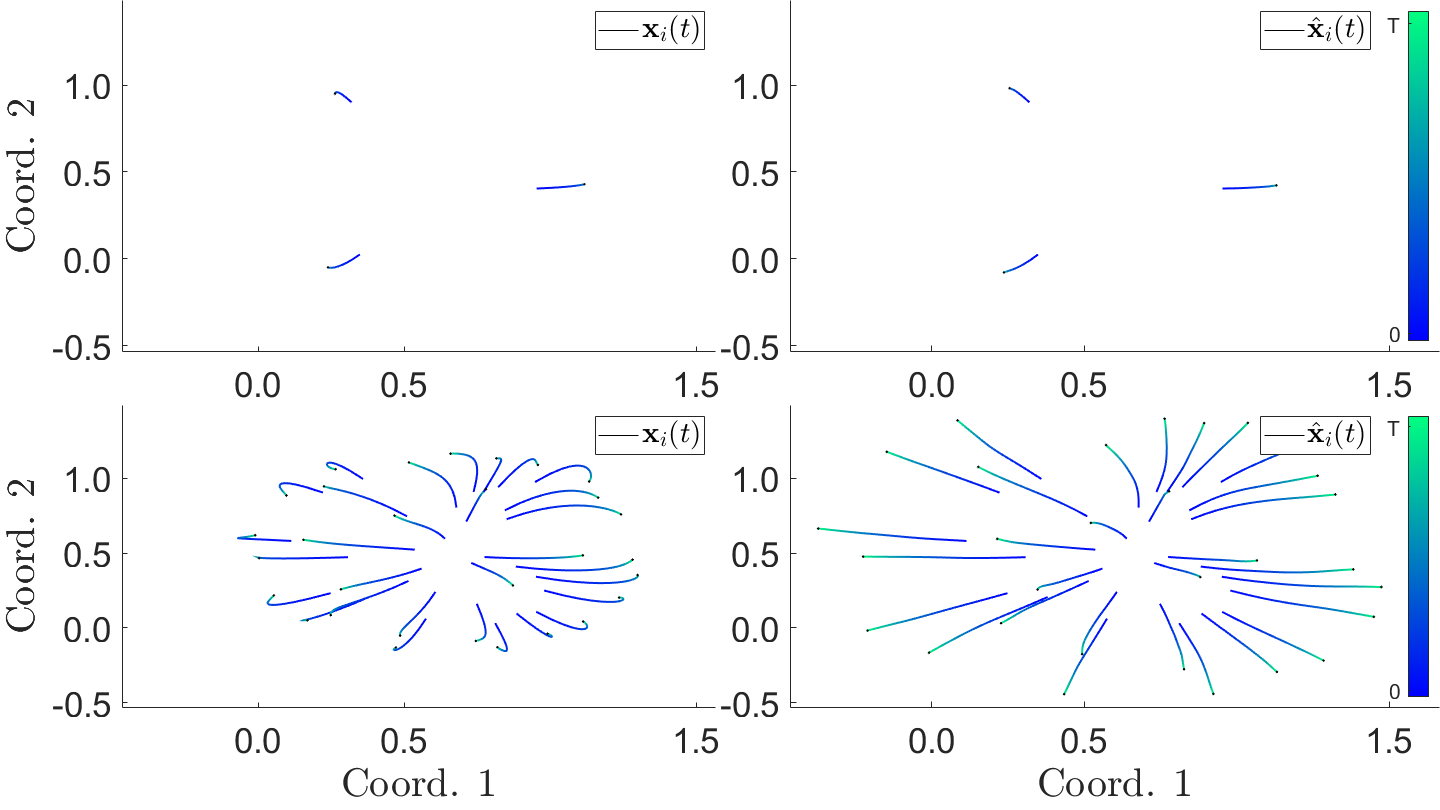}
        \caption{$\bX(t)$ vs $\hat\bX(t)$}
    \end{subfigure}%
    ~ 
    \begin{subfigure}[t]{0.48\textwidth}
        \centering
        \includegraphics[width = 0.9\textwidth]{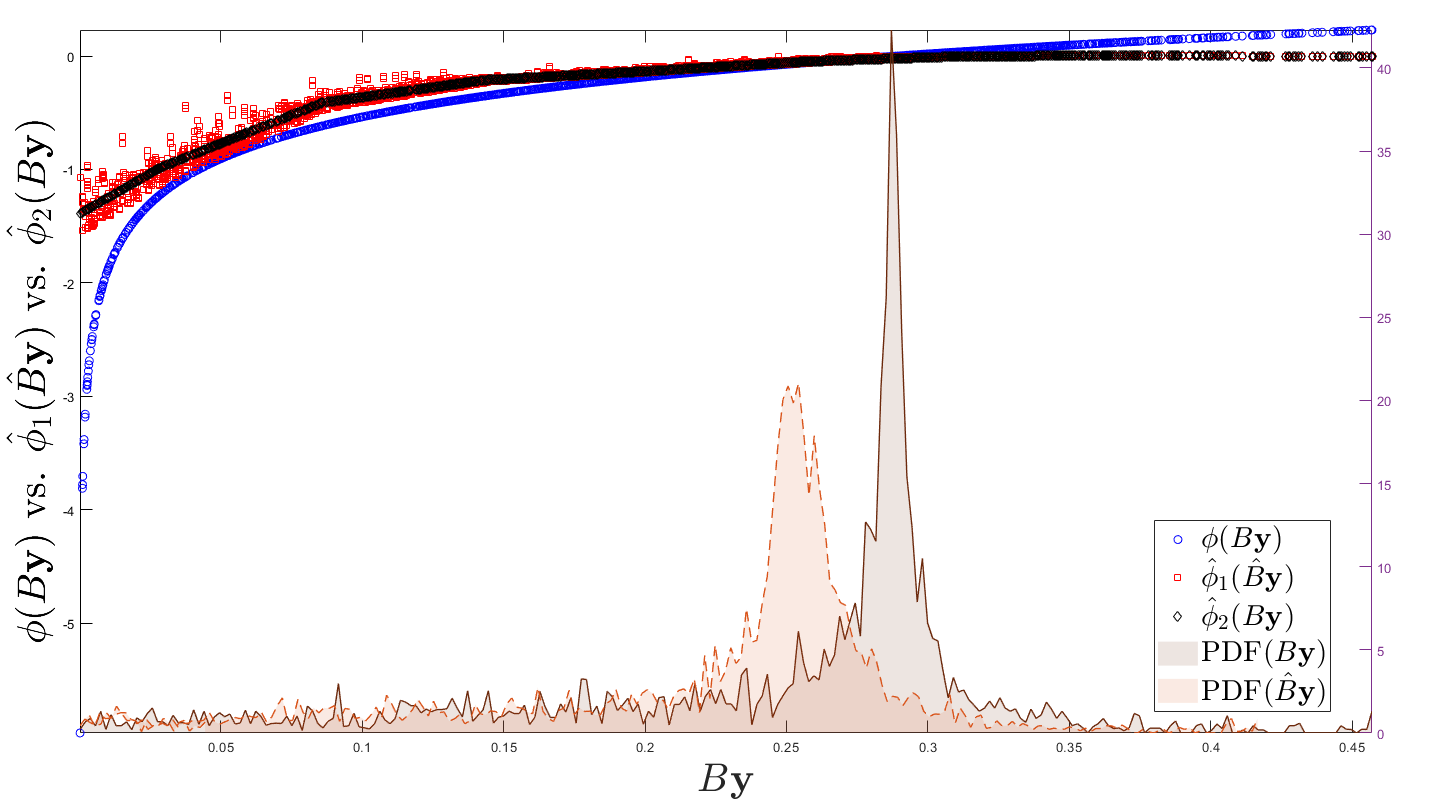}
        \caption{$\phi$ vs $\hat\phi$ (with or without $B$)}
    \end{subfigure}
    \caption{Power Law Dynamics introduced in~\cite{KSUB2011}, with $N = 3$ agents (learned from an updated algorithm), for other parameters see \cite{FENG2022162}.}
\end{figure}
In other words, the interaction kernel $\Phi$ can be factorized (as composition of functions) as a {\bf{reduced interaction kernel}} $\phi:\R^{d'}\rightarrow\R$, and {\bf{reduced variables}} $\bvar:\R^d\times\R^d\rightarrow\R^{d'}$, i.e.
\begin{equation}
\Phi(\bx_i,\bx_{i'})=\phi(\bvar(\bx_i,\bx_{i'})).
\label{e:PhiFactorization}
\end{equation}
When $\bvar(\bx_i,\bx_{i'}):=||\bx_i-\bx_{i'}||$, $\Phi(\bx_i, \bx_{i'}) = \phi(\bvar(\bx_i,\bx_{i'}))=\phi(\norm{\bx_i - \bx_{i'}})$, with $\phi:\R_+\rightarrow\R$, it becomes the simple model which is a special case of what we discussed above. 

Given observations of trajectory data, i.e. $\{\bx_i^{(m)}(t_l), \dot\bx_i^{(m)}(t_l)\}_{i, l, m = 1}^{N, L, M}$, with $0 = t_1 < \cdots < t_L = T$, initial conditions $\bx_i^{(m)}(t_1) \sim \mu_0$ (some unknown probability distribution on the state space $\R^{dN}$) and $m = 1, \ldots, M$, we are interested in estimating the interaction kernel $\Phi$, the only unknown quantity in \eqref{eq:first_order}. Note that this is again not a regression problem, but an inverse problem, since values of the function $\Phi$ are not observed, but only averages of such values, as per the right-hand side of \eqref{eq:first_order}, are.

We proceed by further assuming that $\bvar$ can be factorized as $\bvar(\bx_i,\bx_{i'}):=B\bz(\bx_i,\bx_{i'})$, where $\bz:\R^d\times\R^d\rightarrow\R^D$ is a {\em{known}} feature map from pairs of states to high-dimensional feature vectors, and the feature reduction map $B:\R^D\rightarrow\R^{d'}$ is an {\em{unknown}} linear map to be estimated. Since we will estimate the reduced interaction kernel $\phi$ in a non-parametric fashion, what really matters is the range of $B$, and so we may assume $B$ to be orthogonal. Therefore, the factorization \eqref{e:PhiFactorization} may be rewritten in the form $\Phi(\bx_i, \bx_{i'}) = \phi(B\bz(\bx_i,\bx_{i'}))$. To estimate $\phi$ and $B$ we proceed in two steps.

{\bf{Step $1$. Estimating the feature reduction map $B$, and the variables ${\bvar}$}}.
While the choice of the feature map $\bz$ is typically application-dependent, and can incorporate symmetries, or physical constraints on the system, a rather canonical choice is the map to polynomials in the states, and  here we restrict ourselves to second-order polynomials, and therefore assume that:
\begin{equation}\label{e:yfeatureMap}
\begin{aligned}
&\bz (\bx_i,\bx_{i'}) \\
&\coloneqq \big[ \bx_i , \bx_{i'} , ( (\bx_i)_j(\bx_i)_{j'})_{\substack{j\le j'}}, (( {\bx_{i'})_j}(\bx_{i'})_{j'})_{\substack{j\le j'}}, ((\bx_i)_{j} (\bx_{i'})_{j'})_{\substack{j,j'=1,\dots,d}} \big]^T.
\end{aligned}
\end{equation}
Here $\bz \in \R^{D =2d^2 + 3d}$. Higher-order polynomials can be added for complicated interactions and how to optimally choose such polynomial basis will be for future study.

The techniques of \cite{Martion2021multi}, similar to the previously existing multi-index regression works, are only applicable to the regression setting, and therefore not directly applicable. However, at this point, we note that if we allow the learning procedure to conduct ``experiments'' and have access to a simulator of \eqref{eq:first_order} for different values of $N$, and in particular for $N=2$, it can conduct observations of trajectories of two agents, and from those equations that reveal the values taken by the interaction kernels can be extracted, 
i.e., for $N=2$,
\[
\begin{aligned}
\dot\bx_1 &= \frac{1}{2}\Phi(\bx_1, \bx_2)(\bx_2 - \bx_1), \\
\dot\bx_2 &= \frac{1}{2}\Phi(\bx_2, \bx_1)(\bx_1 - \bx_2), \\
\end{aligned}
\]
we can then transform the equations as follows
\[
\begin{aligned}
\frac{2\inp{\dot\bx_1, \bx_2 - \bx_1}}{\norm{\bx_2 - \bx_1}^2} = \Phi(\bx_1, \bx_2)\,,\, \frac{2\inp{\dot\bx_2, \bx_1 - \bx_2}}{\norm{\bx_1 - \bx_2}^2} = \Phi(\bx_2, \bx_1). \\
\end{aligned}
\]
Consider $\bz_{1,2}:=\bz(\bx_1,\bx_ 2) \in \R^{D}$ defined in \eqref{e:yfeatureMap}, since we assume that $\Phi(\bx_1, \bx_2) = \phi(B\bz_{1, 2})$, we obtain a regression problem:
\[
\phi(B\bz_{1, 2}) = \psi_{1, 2} \coloneqq 2\frac{\inp{\dot\bx_1, \bx_2 - \bx_1}}{\norm{\bx_2 - \bx_1}^2} = \Phi(\bx_1,\bx_2). \\
\]
Similar definitions are used for $\bz_{2, 1}$ and $\psi_{2, 1}$. 

Therefore it takes us back to a regression setting, and we can apply the Multiplicatively Perturbed Least Squares (MPLS) approach of \cite{Martion2021multi} to obtain an estimate $\hat B$ of the feature reduction map $B$. MPLS decomposes the regression function into linear and nonlinear components: $\Phi(\bx_{i}, \bx_{i'}) = \phi(B\bz_{i, i'}) = \inp{\beta, \bz_{i, i'}} + g(A\bz_{i, i'})$, where $(i, i') = \{(1,2), (2, 1)\}$ and $g$ is orthogonal to linear polynomials.  The intrinsic domain of $\Phi$, the row-space of $B$, is hence spanned by $\beta$ and the rows of $A$, which are estimated by an ordinary linear approximation and, respectively, by the top right singular vectors of a matrix of ``slope perturbations''.  Let us re-index the observations as $\{\bz_q\}_{q=1}^Q:=\{\bz_{1, 2}^{(m)}(t_l), \bz_{2, 1}^{(m)}(t_l)\}_{l, m = 1}^{L, M}$, with the corresponding $\{\psi_q\}_{q=1}^Q:=\{\psi_{1, 2}^{(m)}(t_l), \psi_{2, 1}^{(m)}(t_l)\}_{l, m = 1}^{L, M}$, where $Q=2LM$. Then we can estimate $B\in \R^{d' \times  D}$ via the MPLS algorithm as follows:

\textbf{Algorithm [MPLS]}
Inputs: training data $\{\bz_q, \psi_q\}_{q = 1}^Q$, 
partitioned into subsets $\mathcal{S}$ and $\mathcal{S}'$, each of size, $|\mathcal{S}|$ and $|\mathcal{S}'|$, at least $\lfloor \frac{Q}{2}\rfloor$; 
parameters $K$ and $\lambda$, with $K\gtrsim d'\log d'$ and $\lambda \approx \frac{1}{D}$ (choosing $K$ and $\lambda$ is discussed in \cite{Martion2021multi}).

\begin{enumerate}
\item Compute an ordinary least squares linear approximation $\hat \beta$ to $\mathcal{S}'$:
\[ 
\begin{aligned}
\hat\beta \coloneqq \argmin{\beta'\in\mathbb{R}^{D}}\frac{2}{|\mathcal{S}'|} \sum_{(\bz', \psi')\in\mathcal{S}'}\Big(\psi' - \inp{\mathbf{\beta'}, \bz'}\Big)^2.
\end{aligned}
\]
\item Let $\mathcal{R} \coloneqq \Big\{(\bz - \langle \hat\beta, \bz\rangle \|\hat\beta\|^{-2} \hat \beta, \psi - \inp{\hat\beta, \bz}) \big| (\bz, \psi)\in\mathcal{S}\Big\}$ be the residual data of this approximation on $\mathcal{S}$, paired with $\bz$s projected away from $\hat\beta$.
\item Pick $\bu_1,\ldots,\bu_K$ in $\R^{D}$ (e.g., a random subset of $\{\bz_q\}_{q = 1}^Q$). For each $\bu_i$, center the residuals to their weighted mean: with $w(\tilde \bz,\bu_i)=\exp(-\lambda\norm{\tilde \bz - \bu_i}^2)$,
{\small
\[
\widetilde{ \mathcal{R}} \coloneqq \Bigg\{\bigg({\tilde \bz, r - \frac{\sum_{(\tilde \bz, r) \in \mathcal{R}} w(\tilde \bz; \bu_i)r}{\sum_{(\tilde \bz, r) \in \mathcal{R}} w(\tilde \bz;\bu_i)}}\bigg)\bigg| (\tilde \bz, r) \in \mathcal{R}\Bigg\} \,,
\]}
\!\!\!and compute the slope perturbation via least squares:
\[  
\mathbf{\hat p}_i \coloneqq \argmin{\mathbf{p} \in \R^{D}} \frac{2}{|\mathcal{S}|} \sum_{(\tilde \bz, \tilde{r}) \in \widetilde{\mathcal{R}}} (w_k(\tilde \bz, \bu_i)\tilde{r} - \inp{\mathbf{p}, \tilde \bz})^2
\]

\item Let $\hat P\in\R^{K \times D}$ have rows $\mathbf{\hat p}_i$'s, $i=1,\dots,K$, and compute the rank-$d'$ singular value decomposition of $\hat P \approx U_{d'} \Sigma_{d'} V_{d'}^T$. Return $\hat A := V_{d'}^T \in\mathbb{R}^{d'\times D}$ and $\hat\beta$.
\end{enumerate}

While further study on how to optimally add the higher order terms in \eqref{e:yfeatureMap} is ongoing, the MPLS algorithm we applied is not cursed by the dimension $D$, making it possible to increase the dimension of (the range of) the feature map $\bz$ with relatively small additional sampling requirements.

{\bf{Step $2$. Estimating the reduced interaction kernel $\phi$}}.
Once an estimate $\hat B$ for the feature reduction map $B$ has been constructed, we proceed to estimate $\phi$. Always in the case $N=2$, we proceed by projecting the pairs of states, using the estimated feature reduction map $\hat B$, to $\R^{d'}$, and use a non-parametric regression technique on that subspace, aimed at minimizing, over a suitable set of functions $\psi\in\hypspace$, the error functional
\[
\mE_{L, M, \hypspace}(\varphi)\coloneqq\frac{1}{LM}\sum_{l, m = 1}^{L, M}\norm{\dot\bX_l^m - \rhs_{\varphi}(\bX_l^m)}_{\mS}^2, 
\]
where
\[
\rhs_{\varphi}(\bX_l^m) = \begin{bmatrix} \vdots \\ \frac{1}{N}\sum_{i' = 1, i' \neq i}^N\varphi(\hat B\bz_{i, i', l}^m)(\bx_{i'}^m(t_l) - \bx_{i}^m(t_l)) \\ \vdots \end{bmatrix},
\]
where $\bz_{i, i', l}^m = \bz(\bx_{i}^m(t_l), \bx_{i'}^m(t_l))$. We will choose $\hypspace$ to be a convex and compact (in the $L^\infty$ norm) subset of a subspace of functions of the estimated variables $\hat B\by$, e.g. spanned by splines with knots on a grid, or piecewise polynomials. The dimension of $\hypspace$ will be chosen as a suitably increasing function of the number of training trajectories $M$, following the ideas of \cite{lu2019nonparametric} and \cite{miller2023learning} (in the case of multiple reduced variables), see \cite{FENG2022162} for detailed discussion.
\subsection{Coupled Systems}
We explore two types of coupled systems to introduce more intricate sets of states. Initially, we consider the state variable $\by_i$ as $\by_i = \begin{bmatrix} \bx_i \ \zeta_i \end{bmatrix}$, where $\bx_i \in \R^d$ and $\zeta_i \in \R$ (we can also take $\zeta_i \in \R^{d'}$ for any $d' \ge 1$), and thus, $\by_i \in \R^{d + 1}$. Typically, $\bx_i$ describes the position, while $\zeta_i$ denotes the phase, excitation, or opinion (as seen in the swarmalator model, for instance \cite{hao2023}). The interaction arises not only from each $\by_i$ affecting the agents but also from the intra-agent interaction of $\bx_i$ and $\zeta_i$, resulting in a set of more complex patterns. The evolution of $\by_i$ is governed by the following system of ODEs:
\begin{equation}\label{eq:coupled_system_first}
\begin{cases}
    \dot\bx_i &= \force^{\bx}(\by_i) + \sum_{i' = 1, i' \neq i}^N\frac{1}{N_{\clof{i}}}\phi^E_{\clof{i}, \clof{i'}}(\bxi^E(\by_i, \by_{i'}))(\bx_{i'} - \bx_i) \\
    \dot\zeta_i &= \force^{\zeta}(\by_i) + \sum_{i' = 1, i' \neq i}^N\frac{1}{N_{\clof{i}}}\phi^{\zeta}_{\clof{i}, \clof{i'}}(\bxi^{\zeta}(\by_i, \by_{i'}))(\zeta_{i'} - \zeta_i)
\end{cases}, \quad i = 1, \cdots, N.
\end{equation}
Here, $\force^{\bx}: \R^{d + 1} \rightarrow \R^{d}$ represents a force governing how the interaction of the $i^{th}$ agent and its surrounding environment affects the change of $\bx_i$. Similarly, $\force^{\zeta}: \R^{d + 1} \rightarrow \R$ acts on $\zeta_i$. Additionally, $\bxi^E: \R^{2d + 2} \rightarrow \R^{d^E}$ is an energy-based reduced variable, where $1 \le d^E \ll 2d + 2$. The function $\phi^E: \R^{d^E} \rightarrow \R$ is the energy-based reduced interaction kernel. Similarly, $\bxi^{\zeta}: \R^{2d + 2} \rightarrow \R^{d^\zeta}$ is a $\zeta$-based reduced variable, and $\phi^{\zeta}$ is its corresponding reduced interaction kernel. Given the observations ${\by_i^m(t_l), \dot\by_i^m(t_l)}{i, l, m = 1}^{N, L, M}$, we aim to find the set of reduced interaction kernels\footnote{Here we assume that the reduced variables $\bxi^E$ and $\bxi^{\zeta}$ are known to us.} by minimizing the following loss functionals. First, for $\bvarphi^E = \{\varphi^E_{k_1, k_2} \}_{k_1, k_2 = 1}^K$, we minimize:
\[
\mE^{\bx}_{L, M, \bhypspace^E}(\bvarphi^E) = \frac{1}{LM}\sum_{l, m = 1}^{L, M}\norm{\dot\bY_l^m - \rhs_{\bvarphi^E}(\bY_l^m)}_{\mS'}^2
\]
for $\bvarphi^E = \{\varphi^E_{k_1, k_2} \in \hypspace^E_{k_1, k_2}\}_{k_1, k_2 = 1}^K$ and $\bhypspace^E = \oplus_{k_1, k_2 = 1}^K\hypspace^E_{k_1, k_2}$, 
\[
\bY_l^m = \begin{bmatrix} \vdots \\ \bx_i^m(t_l) \\ \xi_i^m(t_l) \\ \vdots \end{bmatrix}
\]
and
\[
\rhs_{\bvarphi^E}(\bX_l^m) = \begin{bmatrix} \vdots \\ \force^{\bx}(\by_{i, l}^m) + \sum_{i' = 1, i' \neq i}^N\frac{1}{N_{\clof{i}}}\varphi^E_{\clof{i}, \clof{i'}}(\bxi^E(\by_{i, l}^m, \by_{i', l}^m))(\bx_{i', l}^m - \bx_{i, l}^m) \\\vdots \end{bmatrix}
\]
Here $\bx_{i, l}^m = \bx_i^m(t_l)$ and $\by_{i, l}^m = \by_i^m(t_l)$.  Similarly for learning $\bvarphi^\zeta = \{\varphi^\zeta_{k_1, k_2} \}_{k_1, k_2 = 1}^K$,
\[
\mE^{\zeta}_{L, M, \bhypspace^\zeta}(\bvarphi^\zeta) = \frac{1}{LM}\sum_{l, m = 1}^{L, M}\norm{\dot\bY_l^m - \rhs_{\bvarphi^\zeta}(\bY_l^m)}_{\mS'}^2
\]
for $\bvarphi^\zeta = \{\varphi^\zeta_{k_1, k_2} \in \hypspace^\zeta_{k_1, k_2}\}_{k_1, k_2 = 1}^K$ and $\bhypspace^E = \oplus_{k_1, k_2 = 1}^K\hypspace_{k_1, k_2}^\zeta$ and 
\[
\rhs_{\bvarphi^\zeta}(\bY_l^m) = \begin{bmatrix} \vdots \\ \force^{\zeta}(\by_{i, l}^m) + \sum_{i' = 1, i' \neq i}^N\frac{1}{N_{\clof{i}}}\varphi^\zeta_{\clof{i}, \clof{i'}}(\bxi^\zeta(\by_{i, l}^m, \by_{i', l}^m))(\zeta_{i', l}^m - \zeta_{i, l}^m) \\\vdots \end{bmatrix}
\]
Here $\zeta_{i, l}^m = \zeta_i^m(t_l)$.

Next, we consider the state variable $\by_i$ as $\by_i = \begin{bmatrix} \bx_i \ \bv_i \ \zeta_i \end{bmatrix}$, where $\bx_i, \bv_i \in \R^d$, and $\zeta_i \in \R$. Specifically, we require that $\bv_i = \dot\bx_i$. The change of $\by_i$ is governed by the following second-order ODE system:
\begin{equation}\label{eq:coupled_system_second}
\begin{cases}
    \dot\bx_i &= \bv_i, \\
    m_i\dot\bv_i &= \force^{\bv}(\by_i) + \sum_{i' = 1, i' \neq i}^N\frac{1}{N_{\clof{i}}}\Big[\phi^E_{\clof{i}, \clof{i'}}(\bxi^E(\by_i, \by_{i'}))(\bx_{i'} - \bx_i) \\
    &\quad + \phi^A_{\clof{i}, \clof{i'}}(\bxi^A(\by_i, \by_{i'}))(\bx_{i'} - \bx_i)\Big]\\
    \dot\zeta_i &= \force^{\zeta}(\by_i) + \sum_{i' = 1, i' \neq i}^N\frac{1}{N_{\clof{i}}}\phi^{\zeta}_{\clof{i}, \clof{i'}}(\bxi^{\zeta}(\by_i, \by_{i'}))(\zeta_{i'} - \zeta_i)
\end{cases}, \quad i = 1, \cdots, N.
\end{equation}
Similarly, $\force^{\bv}: \R^{2d + 1} \rightarrow \R^{d}$ is a force that governs how the interaction of the $i^{th}$ agent and its surrounding environment affects the change of $\bv_i$. Furthermore, $\force^{\zeta}:\R^{2d + 1} \rightarrow \R$ acts on $\zeta_i$. Additionally, $\bxi^E: \R^{4d + 2} \rightarrow \R^{d^E}$ represents an energy-based reduced variable (where $1 \le d^E \ll 4d + 2$), and $\phi^E:\R^{d^E} \rightarrow \R$ is the corresponding energy-based reduced interaction kernel. Similarly, $\bxi^A: \R^{4d + 2} \rightarrow \R^{d^A}$ represents an alignment-based reduced variable (where $1 \le d^A \ll 4d + 2$), and $\phi^A:\R^{d^A} \rightarrow \R$ is the corresponding alignment-based reduced interaction kernel. Finally, $\bxi^{\zeta}: \R^{4d + 2} \rightarrow \R^{d^\zeta}$ represents a $\zeta$-based reduced variable, and $\phi^{\zeta}$ is its corresponding reduced interaction kernel.

In order to learn $\bphi^E = \{\phi^E_{k_1, k_2}\}_{k_1, k_2 = 1}^K$ and $\bphi^A = \{\phi^A_{k_1, k_2} \}_{k_1, k_2 = 1}^K$, we use a slightly updated loss functional
\[
\mE^{E + A}_{L, M, \bhypspace^E, \bhypspace^A}(\bvarphi^E, \bvarphi^A) = \frac{1}{LM}\sum_{l, m = 1}^{L, M}\norm{\dot\bY_l^m - \rhs_{\bvarphi^{E}, \bvarphi^{A}}(\bY_l^m)}_{\mS'}^2
\]
where
\[
\rhs_{\bvarphi^{E}, \bvarphi^{A}}(\bY_l^m) = \begin{bmatrix} \vdots \\ \force^{\bv}(\by_{i, l}^m) + \sum_{i' = 1, i' \neq i}^N\frac{1}{N_{\clof{i}}}\Big[\varphi^E_{\clof{i}, \clof{i'}}(\bxi^E(\by_{i, l}^m, \by_{i', l}^m))(\bx_{i', l}^m - \bx_{i, l}^m) \\
+ \varphi^A_{\clof{i}, \clof{i'}}(\bxi^A(\by_{i, l}^m, \by_{i', l}^m))(\bv_{i', l}^m - \bv_{i, l}^m)\Big]\\\vdots \end{bmatrix}
\]
where $\bv_{i, l}^m = \bv_i^m(t_l)$.  We can use a similar loss as in $\mE^{\text{Mul}}_{L, M, \hypspace}$ for $\rhs_{\bvarphi^{E}, \bvarphi^{A}}(\bY_l^m)$.  See \cite{miller2023learning} where feature maps, i.e. $\bxi^E(\by_{i, l}^m, \by_{i', l}^m)$ and $\bxi^A(\by_{i, l}^m, \by_{i', l}^m)$, are known.
\subsection{Learning with Gaussian Priors}
Gaussian process regression (GPR) serves as a non-parametric Bayesian machine learning technique designed for supervised learning, equipped with an inherent framework for quantifying uncertainty. Consequently, GPR has found application in the study of ordinary differential equations (ODEs), stochastic differential equations (SDEs), and partial differential equations (PDEs) \cite{heinonen2018learning,archambeau2007gaussian,yildiz2018learning,zhao2020state,raissi2017machine,chen2020gaussian,wang2021explicit,chen2021solving,lee2020coarse,akian2022learning,darcy2021learning}, resulting in more accurate and robust models for dynamical systems. Given the unique characteristics of dynamical data, it necessitates novel concepts and substantial efforts tailored to specific types of dynamical systems and data regimes. In our work, we model latent interaction kernels as Gaussian processes, embedding them with the underlying structure of our governing equations, including translation and rotational invariance. This distinguishes our approach from most other works, which model state variables as Gaussian processes.

Despite the challenges involved, recent mathematical advancements have led to the development of a general physical model based on Newton's second law, such as the methods we discussed in section \ref{sec:other_methods}. This model has been demonstrated to capture a wide range of collective behaviors accurately. Specifically, the model describes a system of N agents that interact according to a system of ODEs, where for each agent  $i=1,\cdots,N$:
\begin{eqnarray}\label{eq:2ndOrder}
\begin{aligned}
&m_i\ddot\bx_i = F_i(\bx_i, \dot\bx_i, \mbf{\alpha}_i) +  \sum_{i'=1}^N \frac{1}{N} \Big[\intkernel^E (|\bx_{i'} - \bx_i|)(\bx_{i'} - \bx_i) \\
&\qquad + \intkernela( |{\bx_{i'} - \bx_i}|)(\dot\bx_{i'} - \dot\bx_i)\Big],
\end{aligned}
\end{eqnarray} 
where $m_i \geq 0$ is the mass of the agent $i$; $\ddot\bx_i \in \mathbb{R}^d$ is the acceleration; $\dot\bx_i \in \mathbb{R}^d$ is the velocity; $\bx_i\in \mathbb{R}^d$ is the position of agent $i$; the first term $F_i$ is a parametric function of position and velocities, modeling self-propulsion and frictions of agent $i$ with the environment, and the scalar parameters $\mbf{\alpha}_i$ describing their strength; $|\bx_j-\bx_i|$ is the Euclidean distance; and the 1D functions $\phi^{E}, \phi^{A}:\mathbb{R}^{+}\rightarrow \mathbb{R}$ are called the  \textit{energy and alignment-based radial interaction kernels} respectively. The $\phi^{E}$ term describes the alignment of positions based on the difference of positions;  the $\phi^{A}$ term describes the alignment of velocities based on the difference of velocities. Our primary objective is to infer the interaction kernels $\bintkernel=\{\intkernele, \intkernela \}$ as well as the unknown scalar parameters $\mbf{\alpha}$ and potentially $\mbf{m}$ from the observed trajectory data. Subsequently, we utilize the learned governing equations to make predictions regarding future events or simulate new datasets.

To learn the model given by \eqref{eq:2ndOrder}, we initiate the process by modeling the interaction kernel functions $\intkernele$ and $\intkernela$ with the priors as two independent Gaussian processes
\begin{equation}
\intkernele \sim \mathcal{GP}(0,K_\thetae(r,r')), \qquad
\intkernela \sim \mathcal{GP}(0,K_\thetaa(r,r')),
\end{equation}
where $K_\thetae$, $K_\thetaa$ are covariance functions with hyperparameters $\boldsymbol{\theta} = (\thetae, \thetaa)$.  $\boldsymbol\theta$ can either be chosen by the modeler or tuned via a data-driven procedure discussed later. 

Then given the noisy observational data $\bbY = [\bY^{(1,1)},\dots,\bY^{(M,L)}]^T \in \mathbb{R}^{dNML}$, and $\bbZ = [\bZ^{(1,1)}_{\sigma^2},\dots,\bZ^{(M,L)}_{\sigma^2}]^T \in \mathbb{R}^{dNML}$ where
\begin{equation}
    \bY^{(m,l)} = \bY^{(m)}(t_l):= 
    \begin{bmatrix}\bX(t)\\ \dot\bX(t) 
    \end{bmatrix} \in \mathbb{R}^{2dN},
\end{equation} 
and
\begin{equation}
    \mbf{m}\bZ^{(m,l)}_{\sigma^2} = \mbf{m}\ddot\bX^{(m,l)}_{\sigma^2} := \force_{\mbf{\alpha}}(\bY^{(m,l)}) + \rhsfo_{\bintkernel}(\bY^{(m,l)}) + \epsilon^{(m,l)},
\end{equation}
with $\rhsfo_\bintkernel(\bY(t)) = \rhsfo_{\intkernele, \intkernela}(\bY(t))$ represents the sum of  energy and alignment-based interactions as in  \eqref{eq:2ndOrder}, and $\epsilon^{(m,l)} \sim \mathcal{N}(0, \sigma^2 I_{dN})$ are i.i.d noise, based on the properties of Gaussian processes. With the priors of $\intkernele$, $\intkernela$, we can train
the hyperparameters $\mbf{m}$, $\mbf{\alpha}$, $\mbf{\theta}$ and $\sigma$ by maximizing the probability of the observational data, which is equivalent to minimizing the negative log marginal likelihood (NLML) (see Chapter 4 in \cite{williams2006gaussian})
\begin{align}
    &-\log p(\mbf{m}\bbZ \vert \bbY,\mbf{\alpha},\mbf{\theta},\sigma^2) \notag \\
    &= \frac{1}{2} (\mbf{m}\bbZ - \force_{\mbf{\alpha}}(\bbY))^T(K_{\rhsfo_{\bintkernel}}(\bbY,\bbY;\mbf{\theta}) + \sigma^2I_{dNML})^{-1}(\mbf{m}\bbZ - \force_{\mbf{\alpha}}(\bbY))\notag\\ 
    & \qquad +\frac{1}{2}\log\vert K_{\rhsfo_{\bintkernel}}(\bbY,\bbY;\mbf{\theta})+\sigma^2 I_{dNML}\vert + \frac{dNML}{2} \log 2\pi.
\label{eq:likelihood}
\end{align}
and the posterior/predictive distribution for the interaction kernels $\intkernel^{\mathrm{type}}(r^*)$, $\mathrm{type}=E \text{ or } A$, can be obtained by
\begin{equation}
    p(\intkernel^{\mathrm{type}}(r^\ast)\vert \bbY,\bbZ,r^\ast) \sim \mathcal{N}(\bar{\intkernel}^{\mathrm{type}},var(\bar{\intkernel}^{\mathrm{type}})),
\end{equation}
where
\begin{equation}
    \bar{\intkernel}^{\mathrm{type}} = K_{\intkernel^{\mathrm{type}},\rhsfo_\bintkernel}(r^\ast,\bbY)(K_{\rhsfo_{\bintkernel}}(\bbY,\bbY)+\sigma^2I_{dNML})^{-1}(\mbf{m}\bbZ - \force_{\mbf{\alpha}}(\bbY)),
\label{eq:estimated phi}    
\end{equation}
\begin{equation}
    var(\bar{\intkernel}^{\mathrm{type}}) = K_{\theta^{\mathrm{type}}}(r^\ast,r^\ast) - K_{\intkernel^{\mathrm{type}},\rhsfo_\bintkernel}(r^\ast,\bbY)(K_{\rhsfo_{\bintkernel}}(\bbY,\bbY)+\sigma^2I_{dNML})^{-1}K_{\rhsfo_\bintkernel,\intkernel^{\mathrm{type}}}(\bbY,r^\ast).
    \label{eq:estimated var phi}
\end{equation}
and $\force_{\mbf{\alpha}}(\bbY) = \mathrm{Vec}(\{\force_{\mbf{\alpha}}(\bY^{(m,l)})\}_{m=1,l=1}^{M,L})\in \mathbb{R}^{dNML}$, $K_{\rhsfo_{\bintkernel}} (\bbY,\bbY;\theta)\in \mathbb{R}^{dNML \times dNML}$ is the covariance matrix between $\rhsfo_{\bintkernel}(\bbY)$ and $\rhsfo_{\bintkernel}(\bbY)$, which can be compute elementwisely based on the covariance functions $K_\thetae$, $K_\thetaa$, $K_{\rhsfo_\bintkernel,\intkernel^{\mathrm{type}}}(\bbY, r^*) = K_{\intkernel^{\mathrm{type}},\rhsfo_\intkernel}(r^*,\bbY)^T$ denotes the covariance matrix between $\rhsfo_{\bintkernel}(\bbY)$ and $\intkernel^{\mathrm{type}}(r^*)$.

\begin{figure}[H]
    \centering
    \begin{subfigure}[t]{0.48\textwidth}
        \centering
        \includegraphics[width = 0.9\textwidth]{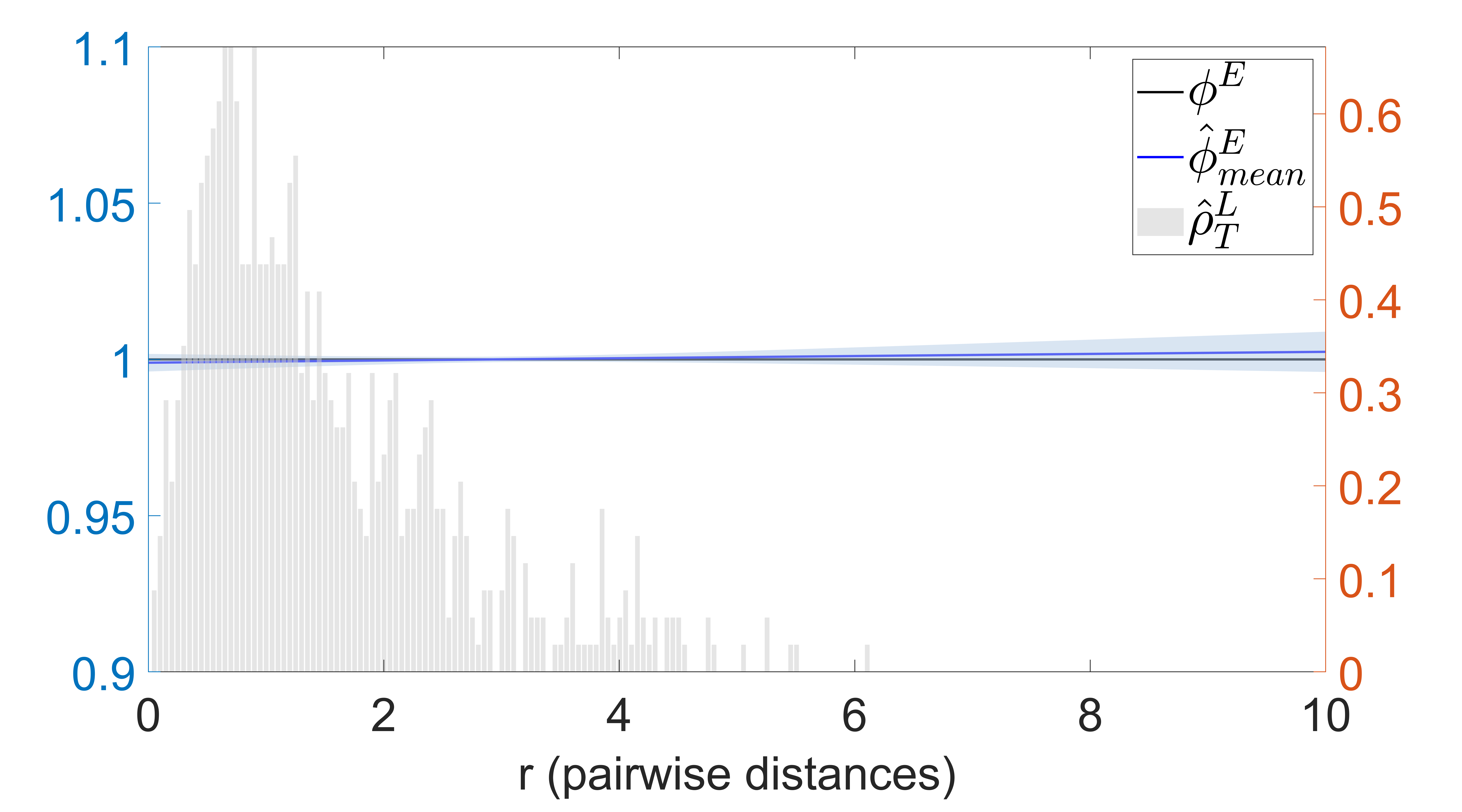}
        \caption{$\phi^E$ vs $\hat\phi^E$}
    \end{subfigure}%
    ~ 
    \begin{subfigure}[t]{0.48\textwidth}
        \centering
        \includegraphics[width = 0.9\textwidth]{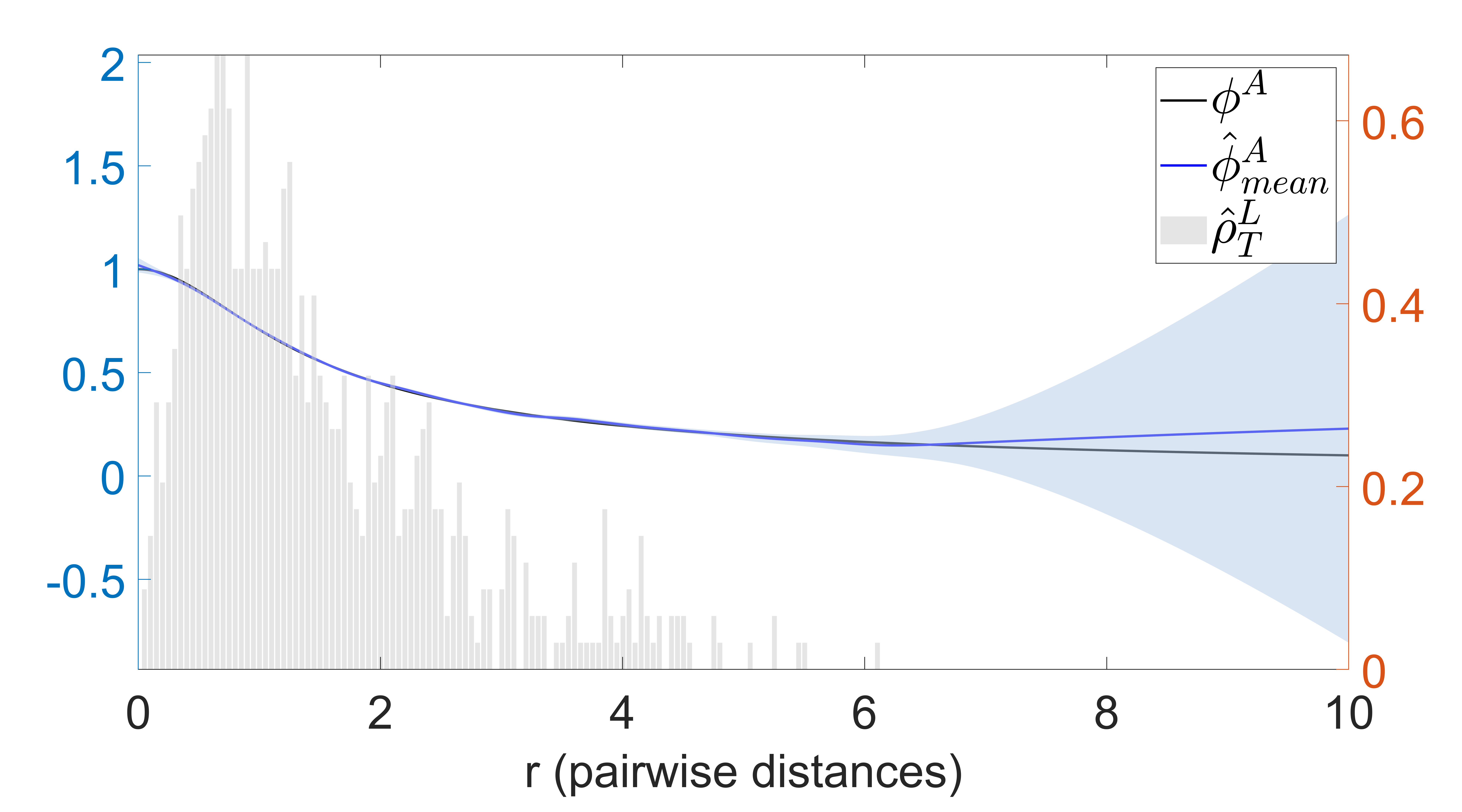}
        \caption{$\phi^A$ vs $\hat\phi^A$}
    \end{subfigure}
    ~ 
    \begin{subfigure}[t]{0.6\textwidth}
        \centering
        \includegraphics[width = 0.9\textwidth]{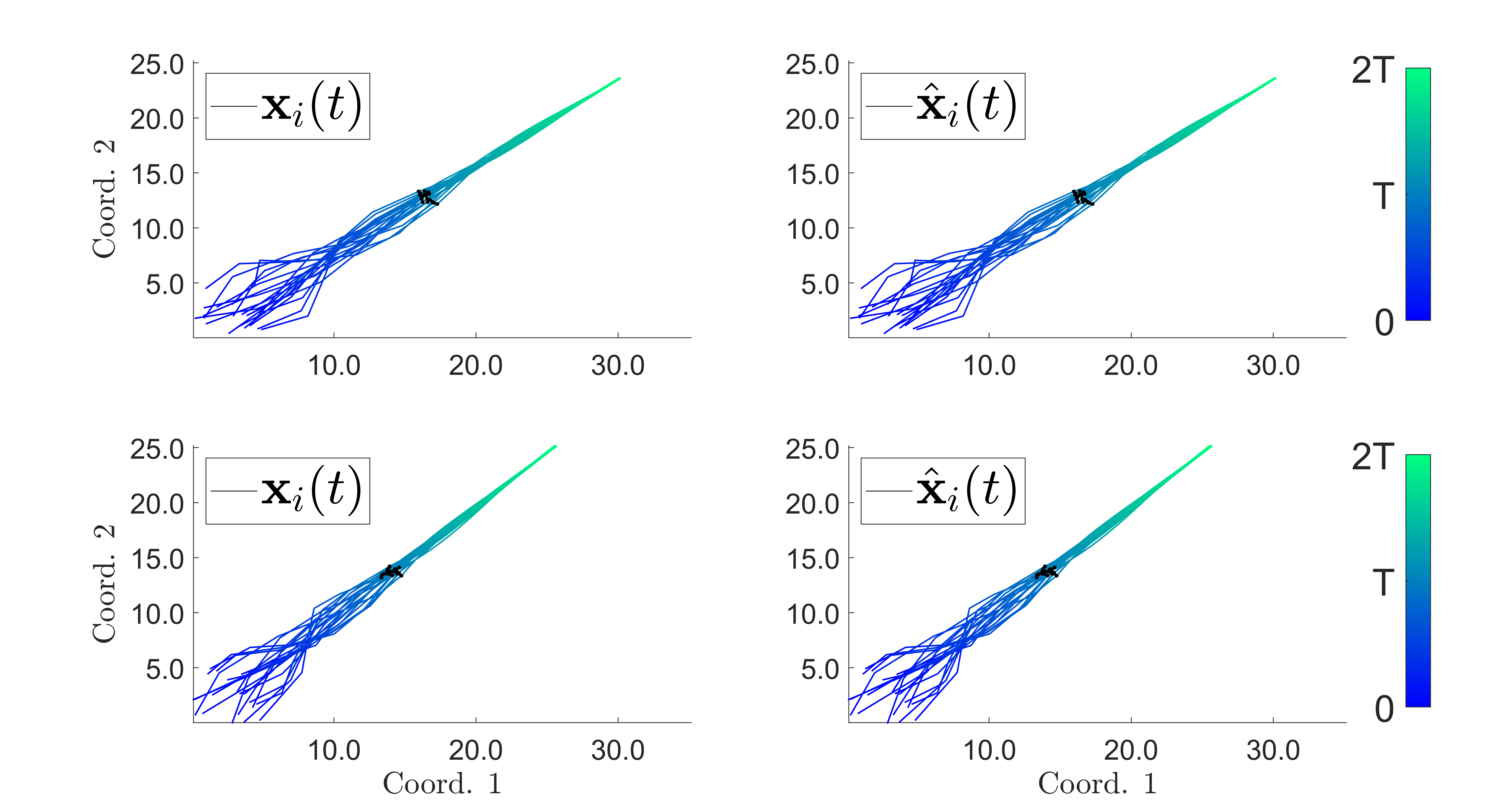}
        \caption{$\bX(t)$ vs $\hat\bX(t)$}
    \end{subfigure}
    \caption{Flocking with external potential (FwEP) model \cite{shu2020flocking} with $\{N,L,M,\sigma\} = \{20,6,3,0.01\}$, and $\phi^E(r) = 1$, $\phi^A(r) = \frac{1}{(1+r^2)^{1/2}}$, for other parameters see \cite{miller2023learning}. The light blue regions are two-standard-deviation bands around the means which indicate the uncertainty of the estimators.}
\label{fig：GP_FwEP}
\end{figure}

Note here the marginal likelihood does not simply favor the models that fit the training data best, but induces an automatic trade-off between data fit and model complexity.  We can use the posterior mean estimators of $\bintkernel$ in trajectory prediction by performing numerical simulations of the equations
\begin{equation}
 \mbf{m}\hat\bZ(t) =\force_{\mbf{\hat\alpha}}(\bY(t)) + \hat\rhsfo_{\bar{\bintkernel}}(\bY(t)).
 \end{equation}
and the posterior variance $var(\bar{\intkernel}^{\mathrm{type}})$ can be
used as a good indicator for the uncertainty of the estimation $\bar{\intkernel}^{\mathrm{type}}$ based on our Bayesian approach, see Figure \ref{fig：GP_FwEP} and also other examples in \cite{feng2022learning,feng2023data}.
In classical regression setting \cite{williams2006gaussian}, there is an interesting link between GP regression with the kernel ridge regression (KRR), where the posterior mean can be viewed as a KRR estimator to solve a regularized least square empirical risk functional.
In our setting, we have noisy functional observations of the interaction kernels, i.e.,   the  $\{r_{\bbX}, r_{\bbV}, \bbZ\}$ instead of the pairs $\{r_{\bbX}, \intkernele(r_{\bbX}), \intkernela(r_{\bbX})\}$, where $r_{\bbX}, r_{\bbV}, \in \mathbb{R}^{MLN^2}$ are the sets contains all the pairwise distances in $\bbX$, and $\bbV$, so we face an inverse problem here, instead of a classical regression problem. Thanks to the linearity of the inverse problem, we can still derive a Representer theorem \cite{owhadi2019operator} that helps understand the role of the hyperparameters: 

\begin{theorem}[Theorem 3.2 in \cite{feng2023data}]  Given the training data $\{\bbY,\bbZ\}$, if the priors $\intkernele \sim \mathcal{GP} (0, \tilde K^E)$, $\intkernela \sim \mathcal{GP} (0, \tilde K^A)$ with $ \tilde K^E=\frac{\sigma^2 K^E} {MNL\lambda^E}$, $ \tilde K^A=\frac{\sigma^2 K^A} {MNL\lambda^A}$ for some $\lambda^E, \lambda^A>0$, then the posterior mean $\bar\bintkernel = (\bar\phi^E,\bar\phi^A)$ in \eqref{eq:estimated phi} coincides with the minimizer, $\bintkernel_{\mHe\times\mHa}^{\lambda,M}$, of the regularized empirical risk functional $\mE^{\lambda,M}(\cdot)$ on $\mHe\times\mHa$ where $\mE^{\lambda,M}(\cdot)$ is defined by
\begin{align}
\label{regularizedrisk1}
\mE^{\lambda,M}(\bintkernelvar):&=\frac{1}{LM}\sum_{l=1,m=1}^{L,M}\| \rhsfo_{\bintkernelvar}(\bY^{(m,l)})-\bZ_{\sigma^2}^{(m,l)}\|^2+\lambda^E \|\intkernelvare\|_{\mHe}^2+\lambda^A \|\intkernelvara\|_{\mHa}^2.
\end{align}
\end{theorem}
From the theorem, it is clear to see how hyperparameters affect the prediction of interaction kernels: $\theta^E$, $\theta^A$, and $\sigma$ jointly affect the choice of Mercer kernels and regularization constant. 

In order to ensure the asymptotic identifiability of the true interaction kernels as the number of observational data snapshots goes to infinity, we study the well-posedness under a statistical inverse problem setting, and provide the coercivity condition in this case:
\begin{definition}[Definition 3.3 in \cite{feng2023data}]\label{coercivityrkhs}
 We say that the system \eqref{eq:2ndOrder} satisfies the coercivity condition if $ \forall \bintkernelvar \in \mHe \times \mHa$,
\begin{align}\label{coercivity}
\|A\bintkernelvar\|^2_{L^2(\rho_{\bY})}=\|\rhsfo_{\bintkernelvar}\|^2_{L^2(\rho_{\bY})}\geq c_{\mHe}\|\intkernelvare\|^2_{L^2( \tilde\rho_{r}^{E})} + c_{\mHa}\|\intkernelvara\|^2_{L^2( \tilde\rho_{r}^{A})}
\end{align}
for some constants $c_{\mHe}, c_{\mHa} > 0$.
\end{definition}
One can prove the well-posedness on a suitable subspace determined by the source conditions on $\intkernele,\intkernela$ with some mild assumptions if the coercivity condition \eqref{coercivity} holds, and both kernels can be recovered with a statistically optimal rate in $M$ under the corresponding RKHS norm:
\begin{equation}
    \|\bintkernel_{\mHe\times\mHa}^{\lambda,M}-\bintkernel\|_{\mHe\times\mHa} \lesssim M^{\frac{-\gamma}{2\gamma+2}},
\end{equation}
where $0<\gamma\leq \frac{1}{2}$, see section 3.2 in \cite{feng2023data} for detailed discussion. A novel operator-theoretical framework is established in \cite{feng2022learning} for the single-kernel systems \eqref{eq:first_order}, which proves the reconstruction error converges at an upper rate in $M$ under  H\"older type source conditions on $\phi$  (Theorem 25 in \cite{feng2022learning}). One can extend this result to the double-kernel case following the same theoretical framework, and we leave it for future investigation. This result generalized the analysis of kernel regression methods \cite{williams2006gaussian} and linear inverse problems to interacting particle systems, and we believe one can obtain more refined rates and bounds using our framework as the bridge in the future.

Compared with the previous works that focused on learning interaction kernels, this method using GPs has the following advantages: (1) it can handle more difficult yet more practical scenarios, i.e., joint inference of scalar parameters $\mbf{\alpha}$ and $\intkernel$, as both are often unknown in practical scenarios. Therefore, our method can learn the governing equations \eqref{eq:2ndOrder}. (2) It provides uncertainty quantification on estimators.  In the ideal data regime, we provide a rigorous analysis and show how it depends on the system parameters. This uncertainty measures the reliability of our estimators, in particular, it can be used to measure the mismatch between our proposed models with the real-world systems.  (3) It has a powerful training procedure to select a data-driven prior and this overcomes the drawback of the previous least square algorithms: there is no criterion to select the optimal choice of function spaces (in terms of both basis and dimensions) for learning so as to minimize the generalization error. See more examples and discussions about the comparisons in \cite{feng2022learning}. 
\section{Conclusion}\label{sec:conclude}
In summary, our paper has presented a thorough exploration of methodologies for deducing the governing structure of collective dynamics from observational data. A detailed examination of the learning framework for first-order models underscores the significance of formulating a variational inverse problem approach based on the intrinsic low-dimensional properties of the dynamical right-hand side. Our methods exhibit efficacy in extending the fundamental approach to encompass multi-species systems, stochastic noise, dynamics constrained on Riemannian manifolds, missing features, coupled systems, and learning with Gaussian priors. Despite the escalating complexity of model equations, once cast into the standard $\dot{\mathbf{Y}} = \rhs(\mathbf{Y})$ form, the formulation of the loss function—the crux of the learning process—becomes evident. We have provided references to our original papers for readers seeking a more in-depth understanding, and additionally, we have conducted comparisons with three other methods, namely SINDy, NeuralODE, and Random Feature Learning.

Numerous avenues warrant consideration for future research. In instances where a single-time observation of the steady state is the sole available data, we have initiated the application of the Reproducing Kernel Hilbert Space approach introduced in \cite{LL2023, lang2020learning} to provide reasonable estimations of $\phi$, up to certain scaling factors. Regarding dynamics with geometric structures, we focus on developing learning methods capable of inferring both the dynamics and the associated Riemannian manifolds. While noise in observation data, encompassing both $\mathbf{Y}$ and $\dot{\mathbf{Y}}$, has been addressed in \cite{lu2019nonparametric}, the scenario of having noisy $\mathbf{Y}$ without information on $\dot{\mathbf{Y}}$ remains an open question. Additionally, exploring the learning of multi-species dynamics without prior knowledge of type information, as introduced in \cite{messenger2022cells}, provides an intriguing starting point. As the number of agents in the system ($N$) grows large, i.e. $N \rightarrow \infty$, computational challenges arise in handling extensive data. Utilizing the corresponding PDEs stemming from the mean-field limit as a guide for learning, as proposed in \cite{LL2023, lang2020learning, messenger2021learning, SHARROCK2023481}, offers promising insights. Partial observation data may render our current learning approach inadequate in capturing the underlying structure; however, we propose a synergistic approach by combining our existing learning methodology with supplementary information derived from the mean-field distribution. The utilization of $\frac{1}{N}$ averaging in the model equations facilitates the derivation of corresponding mean-field PDEs. However, from the perspective of agent-based modeling, such averaging may not align with the actual modeling context. A potential avenue for future exploration involves addressing topological averaging; specifically, considering the right-hand side as $\sum_{j = 1}^N\frac{\phi(\norm{\bx_j - \bx_i})}{\sum_{k = 1}^N\phi(\norm{\bx_k - \bx_i})}$ instead of $\frac{1}{N}\sum_{j = 1}^N\phi(\norm{\bx_j - \bx_i})$ could be a promising direction for further investigation.
\begin{acknowledgement}
JF and MZ made equal contributions to both the research and the composition of the manuscript. They extend their gratitude to Mauro Maggioni from Johns Hopkins University for the invaluable discussions pertaining to this review. Partial support for MZ is acknowledged from NSF-AoF-$2225507$ and the startup fund provided by Illinois Tech.
\end{acknowledgement}
%
%
\bibliographystyle{unsrt}
\bibliography{refs}
\end{document}